%% file: main.tex
\documentclass[10pt,twocolumn,letterpaper]{article}

\usepackage{cvpr}              

\input{preamble}

\definecolor{cvprblue}{rgb}{0.21,0.49,0.74}
\usepackage[pagebackref,breaklinks,colorlinks,citecolor=cvprblue]{hyperref}

\def\paperID{00002} 
\def\confName{CVPR}
\def\confYear{2024}

\usepackage{color}
\usepackage{xcolor}
\usepackage{graphicx}
\usepackage{pifont}
\usepackage{multirow}
\usepackage{xspace}
\usepackage{caption}
\usepackage{amsmath}
\usepackage{enumitem}
\usepackage{wrapfig,lipsum}

\usepackage[linesnumbered, boxed, ruled]{algorithm2e}

\usepackage{colortbl}

\definecolor{cyan}{cmyk}{.3,0,0,0}

\definecolor{springgreen}{RGB}{0,255,0}

\def\ourmethod{{\textit{LocInv}}\xspace}

\newcommand{\model}{\epsilon_\theta}
\newcommand{\modeluncond}{\tilde{\epsilon}_\theta}

\newcommand{\conditioner}{\tau_\theta}
\newcommand{\expec}{\mathbb{E}}
\newcommand{\encoder}{\mathcal{E}}
\newcommand{\decoder}{\mathcal{D}}

\newcommand{\textprompt}{\mathcal{P}}
\newcommand{\textembedding}{\mathcal{C}}

\newcommand{\tokenset}{\mathcal{V}}
\newcommand{\updateset}{{\mathcal{V}_t}}
\newcommand{\updatetoken}{{v_t^k}}
\newcommand{\updateTKsubi}{{v_t^i}}
\newcommand{\updateTKadj}{{a_t^i}}

\newcommand{\crossattn}{\mathcal{A}}

\newcommand{\mlp}{\mathit{l}}

\newcommand{\ddimz}{z}
\newcommand{\denoisez}{{\bar z}}
\newcommand{\interz}{{\tilde z}}

\newcommand{\inputimage}{\mathcal{I}}

\newcommand{\minisection}[1]{\vspace{0.02in} \noindent {\bf #1}\ }

\title{\ourmethod: \textit{Loc}alization-aware \textit{Inv}ersion for Text-Guided Image Editing}

\author{Chuanming Tang\thanks{Equal contributions}\\
University of Chinese Academy of Sciences \\
Computer Vision Center, Spain \\
{\tt\small tangchuanming19@mails.ucas.ac.cn}
\and
Kai Wang\footnotemark[1]\\
Computer Vision Center, Spain\\
{\tt\small kwang@cvc.uab.es}
\and
Fei Yang~\thanks{Corresponding Author}\\
College of Computer Science\\
Nankai University\\
{\tt\small feiyang@nankai.edu.cn}
\and
Joost van de Weijer\\
Computer Vision Center, Spain\\
Universitat Autonoma de Barcelona\\
{\tt\small joost@cvc.uab.es}
}

\begin{document}
\maketitle

\begin{abstract}
Large-scale Text-to-Image (T2I) diffusion models demonstrate significant generation capabilities based on textual prompts.
Based on the T2I diffusion models, text-guided image editing research aims to empower users to manipulate generated images by altering the text prompts. However, existing image editing techniques are prone to editing over unintentional regions that are beyond the intended target area, primarily due to inaccuracies in cross-attention maps.
To address this problem, we propose \textbf{Loc}alization-aware \textbf{Inv}ersion (\ourmethod), which exploits segmentation maps or bounding boxes as extra localization priors to refine the cross-attention maps in the denoising phases of the diffusion process.  
Through the dynamic updating of tokens corresponding to noun words in the textual input, we are compelling the cross-attention maps to closely align with the correct noun and adjective words in the text prompt.
Based on this technique, we achieve fine-grained image editing over particular objects while preventing undesired changes to other regions.
Our method \ourmethod, based on the publicly available \textit{Stable Diffusion}, is extensively evaluated on a subset of the COCO dataset, and consistently obtains superior results both quantitatively and qualitatively. 
\end{abstract}

\begin{figure*}[t]
  \centering
  \vspace{-5mm}
    \includegraphics[width=0.8999\textwidth]{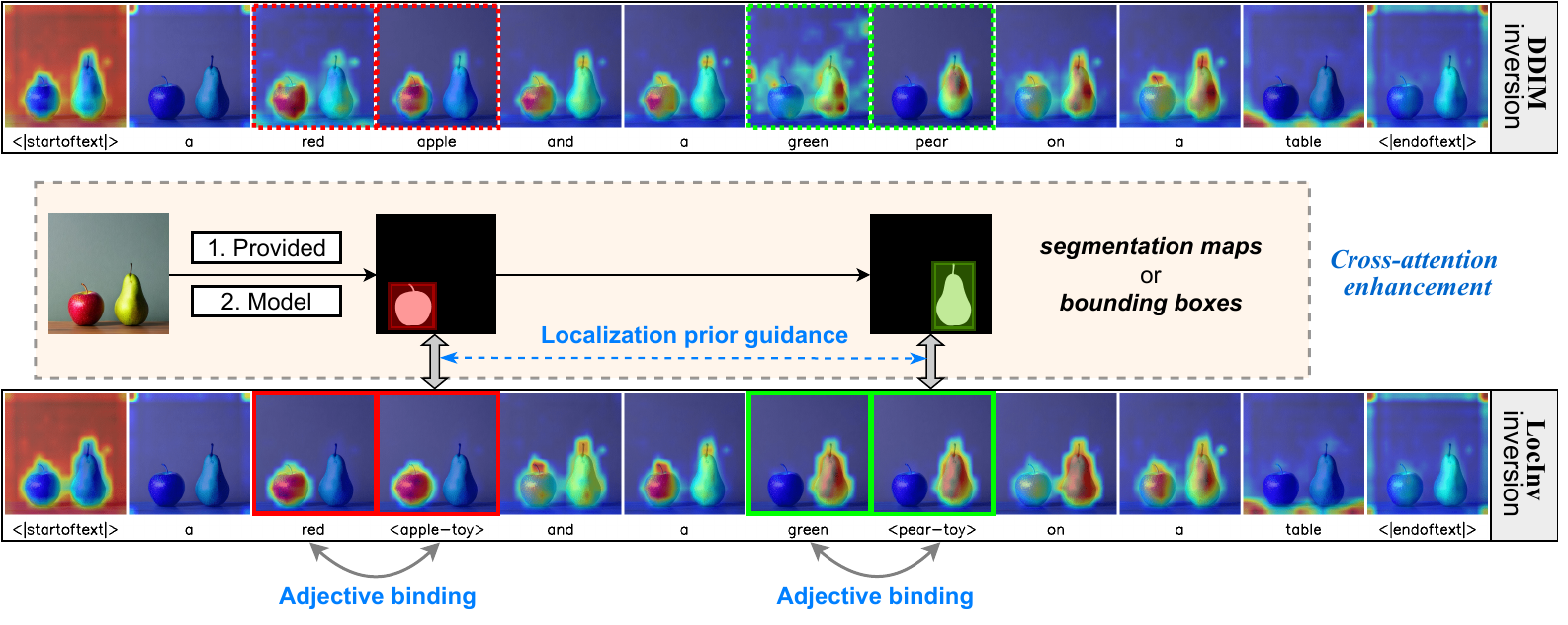}
  \vspace{-4mm}
  \caption{Compared with the naive DDIM inversion, our method \ourmethod aims at enhancing the cross-attention maps by applying localization priors (segmentation maps or detection bounding boxes provided by the datasets or foundation models) to guide the inversion processes. Furthermore, to force strong bindings between adjective and noun words, we constrain the cross-attention similarity between them.}
  \label{fig:cross_attn}
  \vspace{-3mm}
\end{figure*}

\section{Introduction}
\label{sec:intro}

Text-to-Image (T2I) models have made remarkable progress and demonstrated an unprecedented ability to generate diverse and realistic images~\cite{deepfloyd,saharia2022imagen,midjourney,podell2023sdxl,zhou2023shifted}. T2I models are trained on gigantic language-image datasets, necessitating significant computational resources. However, despite their impressive capabilities, they do not directly support \textit{real image editing}, and they typically lack the capability to precisely control specific regions in the image.

Recent research on \textit{text-guided image editing} allows users to manipulate an image using only text prompts~\cite{hertz2023delta_DDS,wang2023mdp,yu2023inpaint_anything,zhang2023forgedit,chen2023fec}. 
In this paper, we focus on text-guided editing, where we aim to change the visual appearance of a specific source object in the image. Several of the existing methods~\cite{mokady2022null,parmar2023zero,tumanyan2022plug,li2023stylediffusion} use DDIM inversion~\cite{song2021ddim} to attain the initial latent code of an image and then apply their proposed editing techniques along the denoising phase. 
Nonetheless, present text-guided editing methods are susceptible to inadvertent alterations of image regions. This arises from the heavy reliance of existing editing techniques on the precision of cross-attention maps. 
DPL~\cite{kai2023DPL} observes the phenomenon that the cross-attention maps from DDIM~\cite{song2021ddim} and NTI~\cite{mokady2022null} do not only correlate with the corresponding objects.
This phenomenon is attributed to \emph{cross-attention leakage}, which is the main factor impeding these image editing methods to work for complex \textit{multi-object} images. To address this, DPL enhances the cross-attention by incorporating additional attention losses.
However, DPL relies on a relatively weak connection between the noun and its associated object. 
This connection occasionally tends to be weak and results in unsatisfactory performance.
Furthermore, given recent advancements in text-based segmentation and detection foundation models~\cite{kirillov2023segment_anything_sam,ShilongLiu2023GroundingDino,xu2022groupvit,zou2023seem}, 
it is now straightforward to automatically obtain strong localization priors into general applications. 

In this paper, we include localization priors to offer enhanced mitigation against \textit{cross-attention leakage}. 
With the introduction of localization priors, our approach, named Localization-aware Inversion (\ourmethod), involves updating the token representations associated with objects at \textit{each timestep}, a technique akin to \textit{dynamic prompt learning}~\cite{kai2023DPL}. In both segmentation and detection scenarios, we optimize two losses—namely, the similarity loss and overlapping loss—to ensure that the cross-attention maps align closely with the provided priors.
Moreover, to accommodate situations in which adjectives describe their associated noun words, we incorporate an additional similarity loss to reinforce the binding between them.
In the experiments, we quantitatively evaluate the quality of cross-attention maps on a dataset \textit{COCO-edit} collected from MS-COCO~\cite{lin2014coco}. 
We further combine \ourmethod with P2P~\cite{hertz2022prompt} to compare with other image editing methods. 
\ourmethod shows superior evaluation metrics  and improved user evaluation. Furthermore, we qualitatively show prompt-editing results for Word-Swap and Attribute-Edit.

\begin{figure*}[t]
  \centering
  \vspace{-3mm}
    \includegraphics[width=0.92222\textwidth]{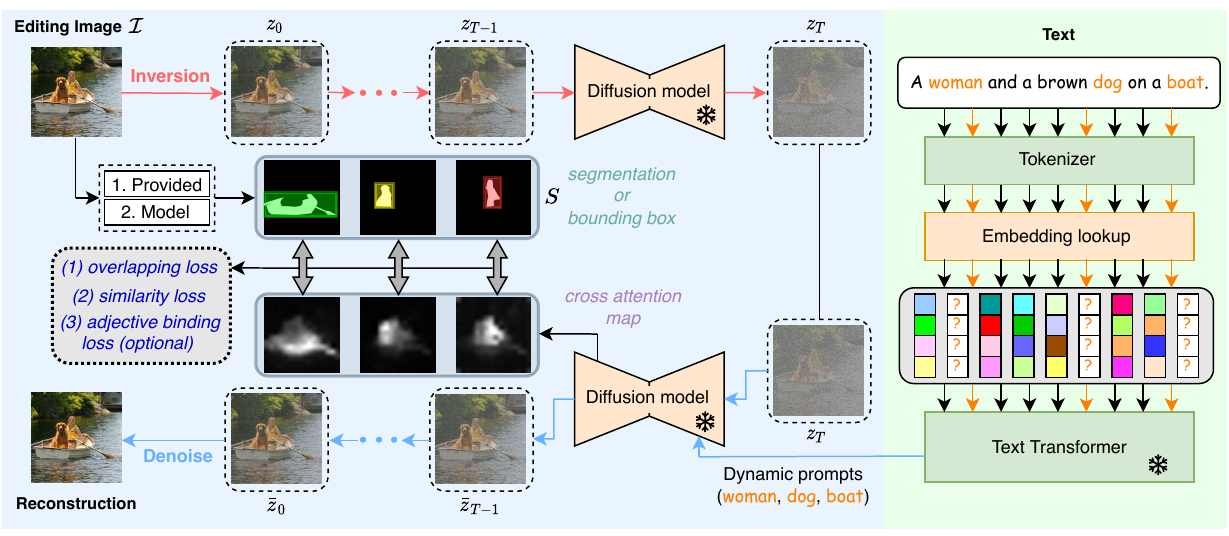}
  \vspace{-3mm}
  \caption{Illustration of our proposed method \ourmethod. The image $\inputimage$ comes with its \textit{localization prior} denoted as $S$ (segmentation maps or detection boxes).
  For each time stamp $t$, the noun (and optionally adjective) words in the text prompt are transformed into dynamic tokens, as introduced in Sec~\ref{subsec:dpl}. 
  In each denoising step $\denoisez_{t-1} \rightarrow \denoisez_t$, we update the dynamic token set $\updateset$ with our proposed overlapping loss, similarity loss and adjective binding loss, in order to ensure high-quality cross-attention maps. 
  }
  \vspace{-5mm}
  \label{fig:framework}
\end{figure*}

\section{Related work}
\label{sec:related_work}

\minisection{Inversion based editing} is mainly relying on the DDIM inversion~\cite{song2021ddim}, which shows potential in editing tasks by deterministically calculating and encoding context information into a latent space and then reconstructing the original image using this latent representation. 
However, DDIM is found lacking for text-guided diffusion models when classifier-free guidance (CFG)~\cite{ho2022classifier} is applied, which is necessary for meaningful editing.
Leveraging optimization on null-text embedding, Null-Text Inversion (NTI)~\cite{mokady2022null} further improved the image reconstruction quality when CFG is applied and retained the rich text-guided editing capabilities of the Stable Diffusion model~\cite{Rombach_2022_CVPR_stablediffusion}. 
Negative-prompt inversion (NPI)~\cite{miyake2023NPI} and ProxNPI~\cite{han2023ProxNPI} reduces the computation cost for the inversion step while generating similarly competitive reconstruction results as Null-text inversion. 
Direct Inversion~\cite{direct_inversion_2023} further enhances the inversion technique by adjusting the editing direction in each timestep to offer essential content preservation and edit fidelity. IterInv~\cite{tang2023iterinv} generalizes the inversion to the DeepFloyd-IF T2I model~\cite{deepfloyd}.

\minisection{Text-guided editing} methods~\cite{bar2022text2live,choi2021ilvr,Kim_2022_CVPR,kwon2023diffusionbased,li2023stylediffusion} of
recent researches~\cite{meng2022sdedit,patashnik2023localizing,huang2023kv} in this topic adopt the large pretrained text-to-image(T2I) models~\cite{chang2023muse,gafni2022make,hong2022sag,ramesh2022dalle2,ramesh2021zero,saharia2022imagen} for controllable image editing.
Among them, Imagic~\cite{kawar2022imagic} and P2P~\cite{hertz2022prompt} attempt structure-preserving editing via Stable Diffusion (SD) models. However, Imagic~\cite{kawar2022imagic} requires fine-tuning the entire model for each image. P2P~\cite{hertz2022prompt} has no need to fine-tune the model and retrains the image structure by assigning \textit{cross-attention} maps from the original image to the edited one in the corresponding text token. InstructPix2Pix~\cite{brooks2022instructpix2pix} is an extension of P2P by allowing human-like instructions for image editing.
NTI~\cite{mokady2022null} further makes the P2P capable of handling real images.
Recently, pix2pix-zero~\cite{parmar2023zero} propose noise regularization and \textit{cross-attention} guidance to retrain the structure of a given image. 
DiffEdit~\cite{couairon2023diffedit} automatically generates a mask highlighting regions of the input image by contrasting predictions conditioned on different text prompts. PnP~\cite{tumanyan2022plug} demonstrated that the image structure can be preserved by manipulating spatial \textit{features} and \textit{self-attention} maps in the T2I models. 

There are also text-guided inpainting methods~\cite{GLIDE,lugmayr2022repaint,grechka2023gradpaint,Rombach_2022_CVPR_stablediffusion} to achieve the editing purposes given user-specific masks.
For example, Blended diffusion~\cite{avrahami2022blended} adapts from a pre-trained unconditional diffusion model and encourages the output to align with the text prompt using the CLIP score. 
Blended latent diffusion (BLD)~\cite{avrahami2023blended} further extend to the LDM~\cite{Rombach_2022_CVPR_stablediffusion}.
Nonetheless, inpainting methods primarily concentrate on filling arbitrary objects in specified regions while ensuring visual coherence with the surrounding areas. These methods do not inherently preserve \textit{semantic similarity} between the source and target objects, as is required for image translation effects.

\minisection{Text-based segmentation and detection models} aim at segmenting or detecting arbitrary classes with the help of language generalization property after pretraining.
One of the most representative prompt-based segmentation model is SAM~\cite{kirillov2023segment_anything_sam}.
Given an image and visual prompt (box, points, text, or mask),
SAM encodes image and prompt embeddings using an image and prompt encoder, respectively which are then combined in a lightweight mask decoder that predicts segmentation masks.
Similar works include  CLIPSeg~\cite{luddecke2022clipseg},  OpenSeg~\cite{ghiasi2022openseg}, GroupViT~\cite{xu2022groupvit}, etc.
For prompt-based object detector, GroundingDINO~\cite{ShilongLiu2023GroundingDino} stands out as the state-of-the-art method by grounding the DINO~\cite{caron2021dino} detector with language pre-training for open-set generalization.
Except that, MaskCLIP~\cite{ding2022maskclip}, X-decoder~\cite{zou2023xdecoder}, UniDetector~\cite{wang2023unidetector} also offer prompt-based detectors.
Leveraging these foundational models, we can acquire localization information as a valuable semantic prior to enhance image inversion. This, in turn, contributes to an overall improvement in image editing performance.

\section{Methodology}
In this section, we provide the description of our method \ourmethod. An illustration of our method is shown in Fig.~\ref{fig:framework} and the pseudo-code in Algorithm~\ref{alg:algorithm}.

\subsection{Preliminary}

\minisection{Latent Diffusion Models.} We use Stable Diffusion v1.4 which is a Latent Diffusion Model (LDM)~\cite{Rombach_2022_CVPR_stablediffusion}. The model is composed of two main components: an autoencoder and a diffusion model. The encoder $\encoder$ from the autoencoder component of the LDMs maps an image $\inputimage$ into a latent code $z_0=\encoder(\inputimage)$ and the decoder reverses the latent code back to the original image as $\decoder(\encoder(\inputimage)) \approx \inputimage$.
The diffusion model can be conditioned on class labels, segmentation masks or textual input. Let $\tau_\theta(y)$ be the conditioning mechanism which maps a condition $y$ into a conditional vector for LDMs, the LDM model is updated by the loss:
\begin{equation}
L_{ldm} = \expec_{z_0 \sim \encoder(x), y, \epsilon \sim \mathcal{N}(0, 1)}\Big[ \Vert \epsilon - \model(z_{t},t, \conditioner(y)) \Vert_{2}^{2}\Big]
\label{eq:ldm_loss}
\end{equation}
The neural backbone $\model$ is typically a conditional UNet~\cite{ronneberger2015unet} which predicts the added noise. More specifically, text-guided diffusion models aim to generate an image from a random noise $z_T$ and a conditional input prompt $\textprompt$. To distinguish from the general conditional notation in LDMs, we itemize the textual condition as $\textembedding=\tau_\theta(\textprompt)$.

\minisection{DDIM inversion.} Inversion aims to find an initial noise $z_T$ reconstructing the input latent code $z_0$ upon sampling. Since we aim at accurate reconstruction of a given image for image editing, we employ the deterministic DDIM sampler:
\begin{equation}
    z_{t+1} = \sqrt{\bar{\alpha}_{t+1}}f_\theta(z_t,t,\textembedding) + \sqrt{1-\bar{\alpha}_{t+1}} \model(z_t,t,\textembedding)
    \label{eq:ddim}
\end{equation}
where $\bar{\alpha}_{t+1}$ is noise scaling factor defined in DDIM~\cite{song2021ddim} and $f_\theta(z_t,t,\textembedding)$ predicts the final denoised latent code $z_0$ as $f_\theta(z_t,t,\textembedding) = \Big[z_t - \sqrt{1-\bar{\alpha}_t} \model(z_t,t,\textembedding) \Big] / {\sqrt{\bar{\alpha}_t}}$.

\minisection{Null-Text inversion (NTI).} To amplify the effect of conditional textual prompts, classifier-free guidance (CFG)~\cite{ho2022classifier} is proposed to extrapolate the conditional noise with an unconditional noise prediction. Let $\varnothing = \tau_\theta(\mathcal{`` "})$ denote the \textit{null-text} embedding, the CFG is defined as:
\begin{equation}
\modeluncond(z_t,t,\textembedding,\varnothing) = w \cdot \model(z_t,t,\textembedding) + (1-w) \ \cdot \model(z_t,t,\varnothing)
\label{eq:classifier_free}
\end{equation}
where we set the guidance scale $w=7.5$ as is standard for LDM~\cite{ho2022classifier,mokady2022null,Rombach_2022_CVPR_stablediffusion}.
However, the introduction of CFG complicates the inversion, and the generated image from the found initial noise $z_T$ deviates from the input image.
NTI~\cite{mokady2022null} proposes a novel optimization which updates the null text embedding $\varnothing_t$ for each DDIM step $t \in [1,T]$ to approximate the DDIM trajectory $\{\ddimz_t\}_0^T$ according to: 
\begin{equation}
    \min_{\varnothing_t} \left \| {\denoisez}_{t-1}- \modeluncond( {\denoisez}_t,t,\textembedding,\varnothing_t) \right \|^2_2
\end{equation}
where $\{\denoisez_t\}_0^T$ is the backward trace from NTI. This allows to edit real images starting from initial noise ${\denoisez}_T=z_T$ using the learned null-text $\varnothing_t$ in combination with P2P~\cite{hertz2022prompt}. 

\subsection{Dynamic Prompt Learning}
\label{subsec:dpl}
Text-based image editing takes an image $\inputimage$ described by an initial prompt $\textprompt$, and aims to modify it according to an altered prompt $\textprompt^*$ in which the user indicates desired changes.
The initial prompt is used to compute the cross-attention maps.
As discussed in Sec.~\ref{sec:intro}, 
 \textit{cross-attention leakage}~\cite{kai2023DPL} is a challenge for existing text-based editing methods, when facing complex scenarios. 
DPL~\cite{kai2023DPL} introduces three losses to enhance the alignment between attention maps and nouns, which rely on the inherent connection between image and prompt and is not always reliable in real-world scenarios. 
In this section, we present our method, denoted as \ourmethod, which leverages localization priors derived from existing segmentation maps (\textit{Segment-Prior}) or detection boxes (\textit{Detection-Prior}). This information can be readily acquired with the assistance of recent advancements in foundation models~\cite{kirillov2023segment_anything_sam,zou2023seem} and has the potential to dramatically strengthen the quality of the cross-attention maps. To simplify, we denote the segmentation map and detection boxes uniformly as $S$.

The cross-attention maps in the Diffusion Model UNet are obtained from $\model(z_t,t,\textembedding)$, which is the first component in Eq.~\ref{eq:classifier_free}. They are computed from the deep features of the noisy image $\psi(z_t)$ which are projected to a query matrix $Q_t=\mlp_Q (\psi(z_t))$, and the textual embedding which is projected to a key matrix $K = \mlp_K (\textembedding)$. Then the attention map is computed as $\mathcal{A}_t=softmax(Q_t \cdot K^T / \sqrt{d})$,
where $d$ is the latent dimension, and the cell $[\mathcal{A}_t]_{ij}$ defines the weight of the $j$-th token on the pixel $i$.
We optimize the word embeddings $v$ corresponding to the initial prompt $\textprompt$ in such a way that the resulting cross-attention $\mathcal{A}_t$ does not suffer from the above-mentioned cross-attention leakage. 
The initial prompt $\textprompt$ contains $K$ noun words and their corresponding learnable tokens at each timestamp  ${\updateset}=\{{v_t^1},...,{v_t^k} ...,{v_t^K}\}$. 
Similar to DPL, \ourmethod updates each specified word in $\updateset$ for each step $t$.
The final sentence embedding $\textembedding_t$ now varies for each timestamp $t$ and is computed by applying the text encoder on the text embeddings.

\subsection{\ourmethod: Localization-aware Inversion}

To update the token representations in each timestep, we propose several losses to optimize the embedding vectors $\updateset$: we develop one loss to address the similarity and another one to ensure high overlapping, between the cross-attention map and its corresponding location prior $S$. 

\minisection{Similarity loss.}
The similarity is defined as the cosine distance between the attention map and the location prior.
\begin{align}
    \mathcal{L}_{sim} = \sum_{\substack{i=1}}^{K}
     \big[1-\mathrm{cos}  (\crossattn_t^{\updateTKsubi},S_t^{\updateTKsubi}) \big ]
\label{eq:cosine_loss}
\end{align}
Nonetheless, our experiments reveal that solely employing the similarity loss leads to lower Intersection over Union (IoU) curves. Given that attention maps are continuous functions, we have additionally introduced an overlapping loss to gently restrict the cross-attention.

\minisection{Overlapping loss.}
This loss is defined as the percentage of the attention map locating in the localization prior as:
\begin{align}
    \mathcal{L}_{ovl} = 1- \frac{\sum_{\substack{i=1}}^{K} \crossattn_t^{\updateTKsubi} \cdot S_t^{\updateTKsubi}}{\sum_{\substack{i=1}}^{K} \crossattn_t^{\updateTKsubi}}
\label{eq:overlap_loss}
\end{align}
By incorporating both losses, our method effectively aligns the cross-attention maps with the localization priors. 
We update the learnable token $\updatetoken$ according to:
\begin{align}
    \underset{\updateset}{\arg\min}  \ \mathcal{L} = \lambda_{sim} \cdot \mathcal{L}_{sim} + \lambda_{ovl} \cdot \mathcal{L}_{ovl} 
\label{eq:total_loss}
\end{align}

\minisection{Gradual Optimization for Token Updates.}
So far, we introduced the losses to learn new dynamic tokens at each timestamp. However, the cross-attention leakage gradually accumulated in the denoising phase. Hence, we enforce all losses to reach a pre-defined threshold at each timestamp $t$ to avoid overfitting the cross-attention maps~\cite{kai2023DPL}.
We express the gradual threshold by an exponential function. For the losses proposed above, the corresponding thresholds at time $t$ are defined as $TH_t = \beta \cdot \exp(-t/\alpha)$. And for each loss we have a group of hyperparameters as $(\beta_{sim},\alpha_{sim}),(\beta_{ovl},\alpha_{ovl}),(\beta_{adj},\alpha_{adj})$.
We verify the effectiveness of this mechanism in our ablation experiments.

\minisection{Null-Text embeddings.}The above described token updating ensures that the cross-attention maps are highly related to the noun words in the text prompt and minimize cross-attention leakage. To  reconstruct the original image, we use NTI~\cite{mokady2022null} in addition to learn a set of null embeddings $\varnothing_t$ for each timestamp $t$. Then  we have a set of learnable word embeddings $\updateset$ and null text $\varnothing_t$ which can accurately localize the objects and also reconstruct the original image.

\subsection{Adjective binding}
\label{sec:adj_bind}
Existing  text-guided image editing methods have focused on translating a source object to a target one. However, often users would like to change the appearance of objects. Typically, in text-guided image editing, this would be done by changing the objects' attributes described by adjectives. However, existing methods fail when editing attributes of the source objects (as shown in Fig.~\ref{fig:attribute_edit}). 
We ascribe this case to the disagreement in cross-attention between the adjective and its corresponding noun word (as evidenced in Fig.~\ref{fig:cross_attn}).

To empower the T2I model with attribute editing capability, we propose to bind the adjective words with their corresponding nouns. To achieve this, we use the Spacy parser~\cite{honnibal2017spacy} to detect the object noun and  adjective words, as so called the adjective-noun pairs $(\updateTKsubi,\updateTKadj)$.
Given these pairs, the adjective binding loss  is defined as the similarity between the attention maps of the adjective and noun words.

\begin{align}
    \mathcal{L}_{adj} = \sum_{\substack{i=1}}^{K}
     \big[1-\mathrm{cos}  (\crossattn_t^{\updateTKsubi},\crossattn_t^{\updateTKadj}) \big]
\label{eq:bind_loss}
\end{align}
This loss ensures maximum overlap between the adjective-noun pairs and is only applied when Adjective-Edit is demanded, and we simply add  $\lambda_{adj} \cdot \mathcal{L}_{adj}$ to Eq.~\ref{eq:total_loss}.

\begin{algorithm}[t]
\SetAlgoLined
\textbf{Input:} A source prompt $\textprompt$, an input image $\inputimage$, the localization prior $S$, $T=50$ \\
\textbf{Output:} A noise vector $\denoisez_T$, a set of updated tokens $\{\updateset\}_1^T$ and null-text embeddings $\{\varnothing_t\}_1^T$ \\
 $\{z_t\}_0^T \gets$DDIM-inv($\inputimage$); \\ 
 Set guidance scale $w=7.5$; \\
 Initialize ${\tokenset}_{T}$ with original noun tokens; \\
  Initialize $\denoisez_T=z_T,\textprompt_T=\textprompt,\varnothing_T = \tau_\theta(\mathcal{``"})$; \\
 \For{$t=T,T-1,\ldots,1$}{
    Initialize ${\textprompt_t}$ by $\updateset$, then ${\textembedding_t}=\tau_\theta({\textprompt_t})$;  \\
    Compute $\mathcal{L}_{sim},\mathcal{L}_{ovl},\mathcal{L}_{adj}$ by Eq.5-8;\\
    \While{$\mathcal{L}_{sim} \geq TH_t^{sim}$  \textbf{or}  $\mathcal{L}_{ovl} \geq TH_t^{ovl}$  \textbf{or}  $\mathcal{L}_{adj} \geq TH_t^{adj}$}{
    $ \mathcal{L} =  \lambda_{sim} \cdot \mathcal{L}_{sim} + \lambda_{ovl} \cdot \mathcal{L}_{ovl} + \lambda_{adj} \cdot \mathcal{L}_{adj}$ \\
    $\updateset \gets \updateset - \nabla_{\updateset}\mathcal{L}$ 
    }
    $\interz_t= \modeluncond({\denoisez}_t,t,{\textembedding_t},\varnothing_t)$ \\
    $\denoisez_{t-1},\varnothing_t \gets NTI(\interz_t,\varnothing_t)$ \\ 
    Initialize $\varnothing_{t-1} \leftarrow \varnothing_t, {\tokenset}_{t-1} \leftarrow \updateset$\\
 }
 \textbf{Return} $\denoisez_T,\{\updateset\}_1^T,\{\varnothing_t\}_1^T$ \\
 \caption{Localization-aware Inversion}
\label{alg:algorithm}
\end{algorithm}

\section{Experiments}
\label{sec:expr}
We demonstrate \ourmethod in various experiments based on the open-source Stable Diffusion~\cite{Rombach_2022_CVPR_stablediffusion} following previous methods~\cite{tumanyan2022plug,parmar2023zero,mokady2022null}. All experiments are done on R6000 GPUs.

\minisection{Datasets.}
For the quantitative ablation study of hyperparameters and partially for the qualitative editing comparison, we select 315 images as a subset \textit{COCO-edit} out of MS-COCO dataset~\cite{lin2014coco}. We compose this subset from various searching prompts (including concepts as airplane, apple, banana, bear, bench, etc.) and store the groundtruth segmentation/detection images for experimental usage. 
Overall, there are 7 search prompts with a single object (noun) in the sentence and 6 with multiple objects.
More detailed information is shown in the supplementary material.

\minisection{Compared methods.}
We organize two groups of methods for qualitative and quantitative comparison. The first group of methods, which are \textit{freezing} the Stable Diffusion models, include NTI~\cite{mokady2022null}, DPL~\cite{kai2023DPL}, pix2pix-zero~\cite{parmar2023zero}, PnP~\cite{tumanyan2022plug}, DiffEdit~\cite{couairon2023diffedit} and MasaCtrl~\cite{cao2023masactrl}.
The second group of methods is \textit{finetuning} the large pretrained T2I model as specific models for image editing, such as SD-inpaint~\cite{Rombach_2022_CVPR_stablediffusion}, InstructPix2Pix~\cite{brooks2022instructpix2pix} and Imagic~\cite{kawar2022imagic}, or taking \textit{masks} as locations for \textit{inpainting}, including SD-inpaint~\cite{Rombach_2022_CVPR_stablediffusion} and BLD~\cite{avrahami2023blended}.

\minisection{Evaluation metrics.}
To quantitatively assess our method's performance, we employ well-established metrics, including LPIPS~\cite{zhang2018lpips}, SSIM~\cite{wang2003ssim}, PSNR,
CLIP-Score~\cite{hessel2021clipscore} and DINO-Sim~\cite{tumanyan2022structure}, to evaluate the edited full image. Additionally, to illustrate the quality of background preservation, we follow DirectInversion~\cite{direct_inversion_2023} to compute LPIPS, SSIM, PSNR, and MSE metrics for regions outside the mask.

\begin{table}[t]
    \centering
    \vspace{-3.5mm}
    \resizebox{0.95\linewidth}{!}{
    \begin{tabular}{c|cccccc}
      \multicolumn{7}{c}{User Study (\%)} \\
      \toprule
      \rotatebox{32}{method}	& \rotatebox{30}{\ourmethod (Seg)}  &	\rotatebox{32}{DiffEdit}	& \rotatebox{32}{MasaCtrl}	& \rotatebox{32}{NTI} & \rotatebox{32}{pix2pix} 	& \rotatebox{32}{PnP} \\
      \midrule
      Edit quality & \textbf{40.0}	& 27.0	& 3.5	& 25.5	& 0.75	& 3.25 \\
      Background & \textbf{25.8}	& 3.7	& 4.5	& 20.0	& 22.7	& 23.3 \\
      \bottomrule
    \end{tabular}
    }
    \vspace{-3.5mm}
    \caption{User study compared with methods freezing the Stable Diffusion~\cite{Rombach_2022_CVPR_stablediffusion}. We request the respondents to evaluate methods in both editing quality and background preservations. }
    \label{tab:user_study}
    \vspace{-7mm}
\end{table}

\subsection{Ablation study}
For the ablation study, we experiment on the \textit{COCO-edit} dataset. To quantitatively assess the localization performance of \ourmethod, we vary the threshold from 0.0 to 1.0 to obtain the segmentation mask from the cross-attention maps. We then calculated the Intersection over Union (IoU) metric using the segmentation ground truth for comparison. Our method can operate with both segmentation maps  and detection bounding boxes as localization priors. Here, we consider the hyperparameters for both these cases. 

\begin{figure*}[t]
  \centering
    \includegraphics[width=0.999\textwidth]{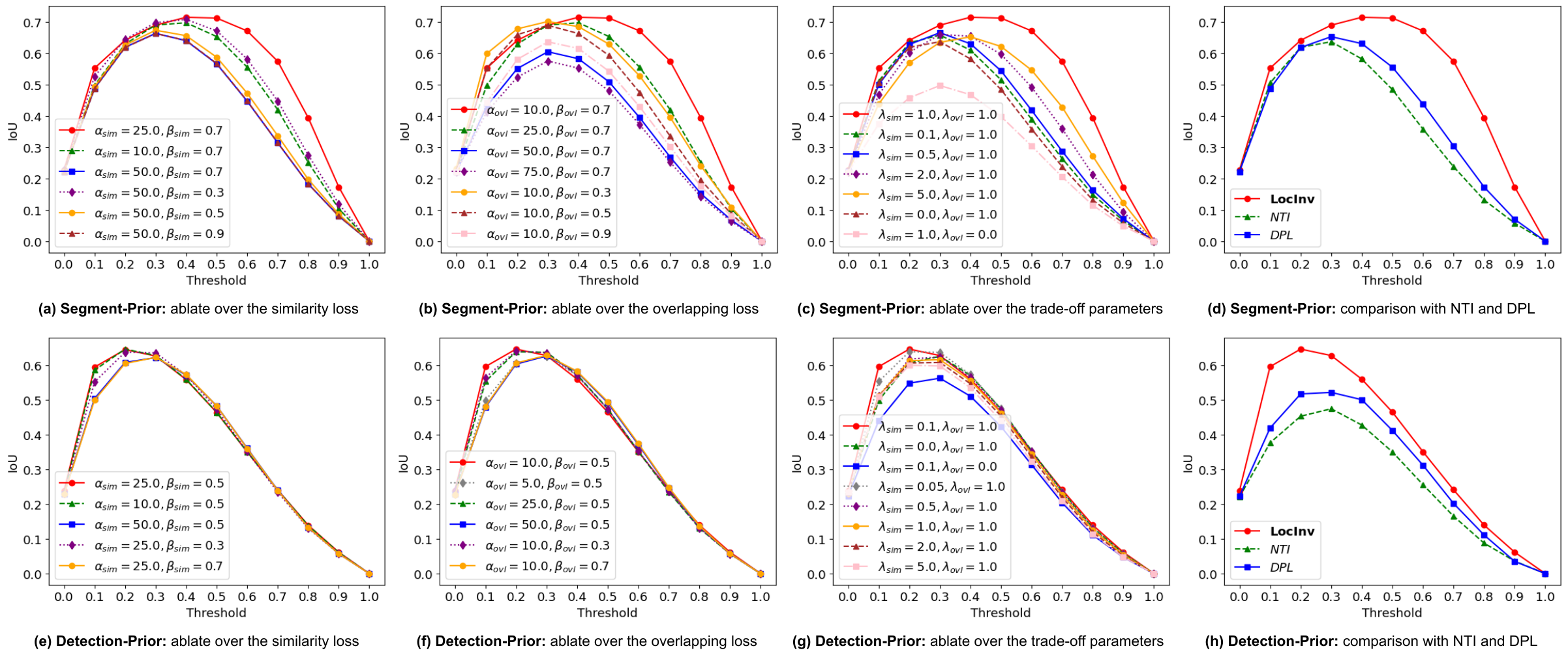}
    \vspace{-4mm}
  \caption{Ablation study over hyperparameters given the Segment-Prior (first row) or Detection-Prior (second row). For the first and second columns, we ablate hyperparameters for the similarity loss and overlapping loss, respectively. Then we illustrate how the trade-off parameters influence in the third column. Lastly, we show the IoU curves of \ourmethod together with NTI and DPL as baseline comparisons. }
  \label{fig:ablation}
    \vspace{2mm}
	\resizebox{0.999\textwidth}{!}{%
        \begin{tabular}{|c | c | cccc | cccc | c|c|}
        \toprule
        & \multirow{2}{*}{Method} & \multicolumn{4}{c|}{Full Image Evaluation} & \multicolumn{4}{c|}{Background Preservation} & \multicolumn{2}{c|}{CLIP-Score($\uparrow$)}  \\
        \cline{3-12}
        & & LPIPS($\downarrow$)	& SSIM($\uparrow$)	& PSNR($\uparrow$) & DINO-Sim($\downarrow$) & MSE($\downarrow$) & LPIPS($\downarrow$) & SSIM($\uparrow$) & PSNR($\uparrow$)				& Edited	& Original \\
        \midrule
        \multicolumn{11}{c}{Multi-Object Image Editing (6 tasks)} \\
        \midrule
        \multirow{7}{*}{\rotatebox[origin=c|]{90}{Freezing SD}} & DiffEdit~\cite{couairon2023diffedit} & 0.2821 & 0.5910 & 18.5804 & \underline{\textit{0.0219}} & 0.0066 & 0.1981 & 0.6888 & 22.7726 & 21.4439 & \multirow{12}{*}{\rotatebox[origin=|c|]{90}{22.05}} \\
        &pix2pix~\cite{parmar2023zero} & 0.4047 & 0.5492 & 19.7397 & 0.0549 & 0.0152 & 0.3347 & 0.6262 & 20.9542 & 21.8384 &\\
        &NTI~\cite{mokady2022null} & 0.2936 & 0.5919 & \underline{\textit{21.7963}} & 0.0433 & 0.0118 & 0.2413 & 0.6644 & 23.1352 & 21.7216 &\\
        & DPL~\cite{kai2023DPL} & \underline{\textit{0.2686}} &	0.6121	&21.3193	&0.0223	&0.0071	&0.2299	&0.6601	&22.2695 &	21.5982 & \\
        &PnP~\cite{tumanyan2022plug} & 0.3960 & 0.5568 & 18.8198 & 0.0384 & 0.0113 & 0.3331 & 0.6243 & 19.8573 & \underline{\textit{21.8470}} &\\
        &MasaCtrl~\cite{cao2023masactrl} & 0.4406 & 0.4671 & 17.3948 & 0.0611 & 0.0198 & 0.3784 & 0.5309 & 18.2041 & 21.7835 &\\
        &\ourmethod (Det) & 0.2935 & 0.5956 & 21.3116 & 0.0272 & 0.0065 & 0.2458 & 0.6532 & 22.5126 & 21.6615 &\\
        &\ourmethod (Seg) & \textbf{0.2523} & \underline{\textit{0.6161}} & \textbf{22.3027} & \textbf{0.0181} & \underline{\textit{0.0054}} & 0.1970 & 0.6905 & 24.3783 & 21.7757 &\\
        \cline{1-11}
        \multirow{6}{*}{\rotatebox[origin=c]{90}{Finetuning SD}} &Imagic~\cite{kawar2022imagic} &  0.7347 & 0.2098 & 9.9586  & 0.1217 & 0.0935 & 0.6166 & 0.3280 & 10.7490 & 21.7566 &\\
        &InstructP2P~\cite{brooks2022instructpix2pix} & 0.3330 & 0.5428 & 17.4007 & 0.0274 & 0.0150 & 0.2462 & 0.6407 & 20.2072 & 21.6666 &\\
        &Inpaint (Det)~\cite{Rombach_2022_CVPR_stablediffusion} & 0.3710 & 0.4853 & 16.9441 & 0.1242 & 0.0398 & 0.2755 & 0.6075 & 21.5161 & \textbf{21.8475} &\\
        &Inpaint (Seg)~\cite{Rombach_2022_CVPR_stablediffusion} & 0.2703  & 0.6040 & 19.2707 & 0.0299 & 0.0061 & \textbf{0.1620} & \underline{\textit{0.7233}} & \underline{\textit{26.3149}} & 21.8315 &\\
        &BLD (Det)~\cite{avrahami2023blended} &  0.3412 & 0.5604 & 17.0294 & 0.0405 & 0.0112 & 0.2424 & 0.6814 & 21.0436 & 21.7218 &\\
        &BLD (Seg)~\cite{avrahami2023blended} &  0.2924 & \textbf{0.6257} & 18.9036 & 0.0258 & \textbf{0.0031 }& \underline{\textit{0.1845}} & \textbf{0.7426 }& \textbf{26.5964 }& 21.7806 &\\
        \midrule
        \multicolumn{11}{c}{Single-Object Image Editing (7 tasks)} \\
        \midrule
        \multirow{7}{*}{\rotatebox[origin=c]{90}{Freezing SD}} & DiffEdit~\cite{couairon2023diffedit} & 0.2990 & 0.5701 & 17.7486 & {{0.0278}} & 0.0057 & 0.1873 & {{0.7136}} & 23.4800 & 21.4608 &  \multirow{12}{*}{\rotatebox[origin=c]{90}{21.77}} \\
        &pix2pix~\cite{parmar2023zero} & 0.4398 & 0.4824 & 17.2601 & 0.0645 & 0.0171 & 0.3180 & 0.6332 & 20.1122 & \underline{\textit{21.8336}}  &  \\
        &NTI~\cite{mokady2022null} & {{0.2758}} & 0.5864 & \textbf{21.4369} & 0.0280 & 0.0092 & 0.1936 & 0.7001 & 24.3014 & 21.7665 &  \\
        & DPL~\cite{kai2023DPL} & \underline{\textit{0.2743}}&	\underline{\textit{0.5906}}	&21.1188	&0.0212	&0.0061	&0.1791	&0.7133	&25.1545	&21.7944 & \\
        &PnP~\cite{tumanyan2022plug} & 0.3983 & 0.5379 & 18.0061 & 0.0338 & 0.0111 & 0.2893 & 0.6688 & 20.0125 & 21.7214 &\\
        &MasaCtrl~\cite{cao2023masactrl} & 0.4004 & 0.4472 & 17.2875 & 0.0430 & 0.0138 & 0.2879 & 0.5957 & 19.4637 & 22.0493  &  \\
        &\ourmethod (Det) & {0.2756} & {0.5867} & 21.1920 & \underline{\textit{0.0196}} & 0.0049 & {{0.1810}} & 0.7118 & 25.1956 & 21.7308  & \\
        &\ourmethod (Seg) &  \textbf{0.2662} & \textbf{0.5952} & \underline{\textit{21.2287}} & \textbf{0.0180} & 0.0047 & 0.1730 & 0.7193 & 25.2118 &  21.8069 &\\
        \cline{1-11}
        \multirow{6}{*}{\rotatebox[origin=c]{90}{Finetuning SD}} &Imagic~\cite{kawar2022imagic} & 0.6657 & 0.2429 & 10.8554 & 0.1418 & 0.0694 & 0.5107 & 0.4313 & 12.3759 & \textbf{21.8340} & \\
        &InstructP2P~\cite{brooks2022instructpix2pix} & 0.3684 & 0.4925 & 16.3615 & 0.0421 & 0.0177 & 0.2519 & 0.6477 & 19.7768 & 21.7257 &   \\
        &Inpaint (Det)~\cite{Rombach_2022_CVPR_stablediffusion} & 0.3034 & 0.5458 & 17.4352 & 0.0301 & \underline{\textit{0.0039}} & {\textbf{0.1570}} & \underline{\textit{0.7300}} & \underline{\textit{26.4033}} & 21.7324 & \\
        &Inpaint (Seg)~\cite{Rombach_2022_CVPR_stablediffusion} & 0.3080 & 0.5380 & 17.3868 & 0.0304 & \textbf{0.0038} & \underline{\textit{0.1599}} & {0.7241} & \textbf{26.4464} & 21.7967 &  \\
        &BLD (Det)~\cite{avrahami2023blended} & 0.4112 & 0.4729 & 15.3787 & 0.0861 & 0.0175 & 0.2676 & 0.6649 & 18.9651 & 21.6654 &\\
        &BLD (Seg)~\cite{avrahami2023blended} & 0.3423 & 0.5571 & 17.4153 & 0.0455 & 0.0040 & 0.1895 & \textbf{0.7478} & 25.6214 & 21.6880 &\\
        \bottomrule
        \end{tabular}
	}
    \vspace{-2mm}
	\captionof{table}{Comparison with various text-based image editing methods based on the evaluation metrics over the \emph{COCO-edit} dataset. We evaluate on single-object and multi-object images editing tasks separately.
    The comparison methods are organized into two groups as we stated in Sec.~\ref{sec:expr}
    The “Seg” and “Det” in the bracket represent the Segment-prior and Detection-Prior, respectively.} 
	\label{tab:eval_metric}
    \vspace{-2mm}
\end{figure*}

In Fig.~\ref{fig:ablation}, we conduct ablation studies over the similarity loss and the overlapping loss.
From Fig.~\ref{fig:ablation}-(c)(g), we observe that only applying one of these losses does not ensure a satisfactory performance. 
Empirically, we find the optimal hyperparameters for the segmentation and detection prior as $\lambda_{sim}=1.0, \alpha_{sim}=50.0, \beta_{sim}=0.7,\lambda_{ovl}=1.0, \alpha_{ovl}=10.0, \beta_{ovl}=0.7$ and $\lambda_{sim}=0.1, \alpha_{sim}=25.0, \beta_{sim}=0.5,\lambda_{ovl}=1.0, \alpha_{ovl}=25.0, \beta_{ovl}=0.3$, respectively. For the adjective binding loss, since there are no sufficient image-text pairs for the ablation study, we empirically set the hyperparameters to be $\lambda_{adj}=2.0, \alpha_{adj}=50.0, \beta_{adj}=0.1$. All results in this paper are generated with this hyperparameter setting.

\begin{figure*}[t]
  \centering
    \includegraphics[width=0.99\textwidth]{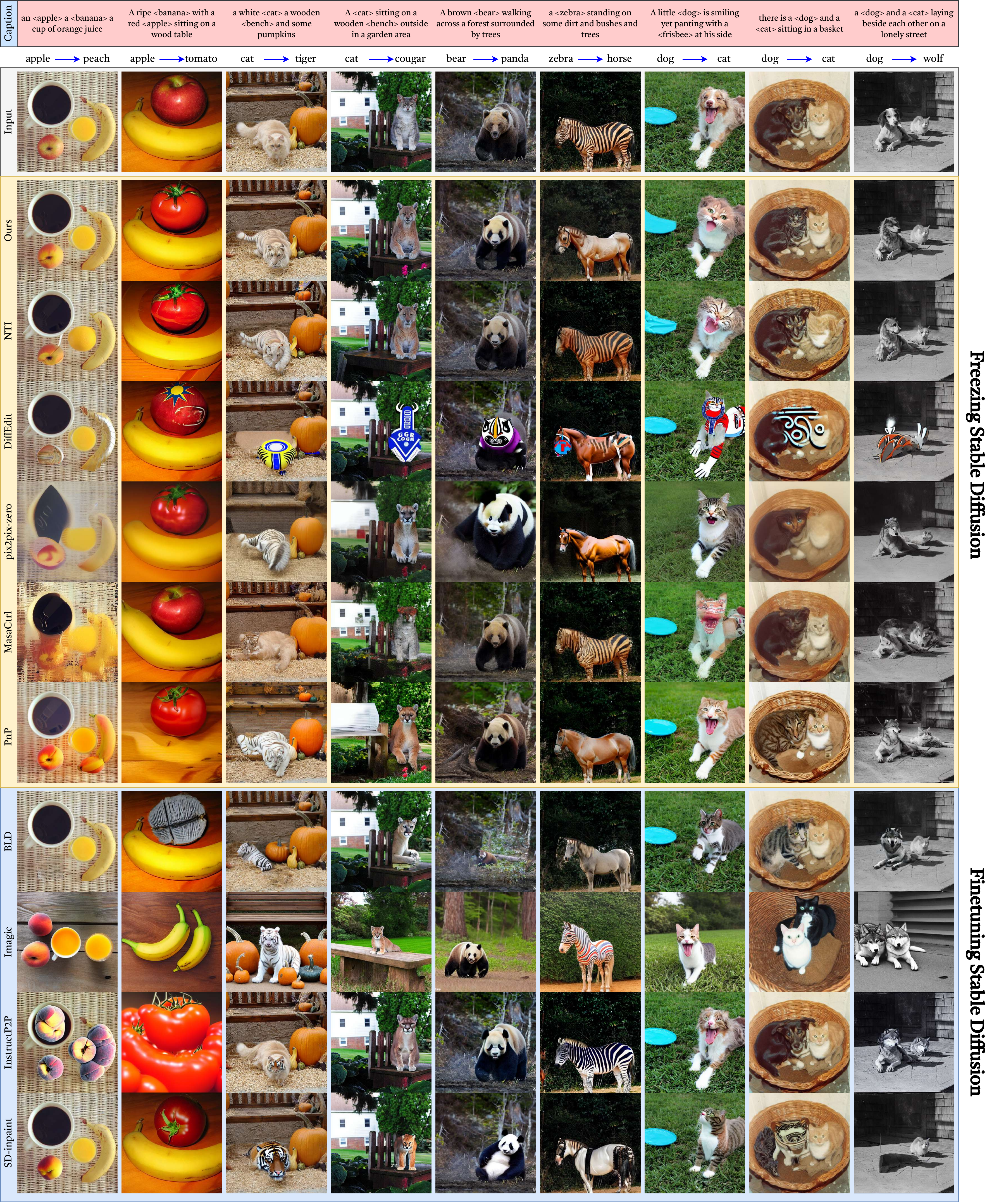}
  \caption{Comparison over the local object Word-Swap editing given the \textit{Segment-Prior}. All examples are from the COCO-edit dataset. We distinguish these comparison methods by (1) freezing the SD~\cite{Rombach_2022_CVPR_stablediffusion} models; (2) fine-tuning the SD models or mask-based inpainting.}
  \label{fig:local_swap_comp}
\end{figure*}

\begin{figure*}[t]
  \centering
  \vspace{-2mm}
    \includegraphics[width=0.999\textwidth]{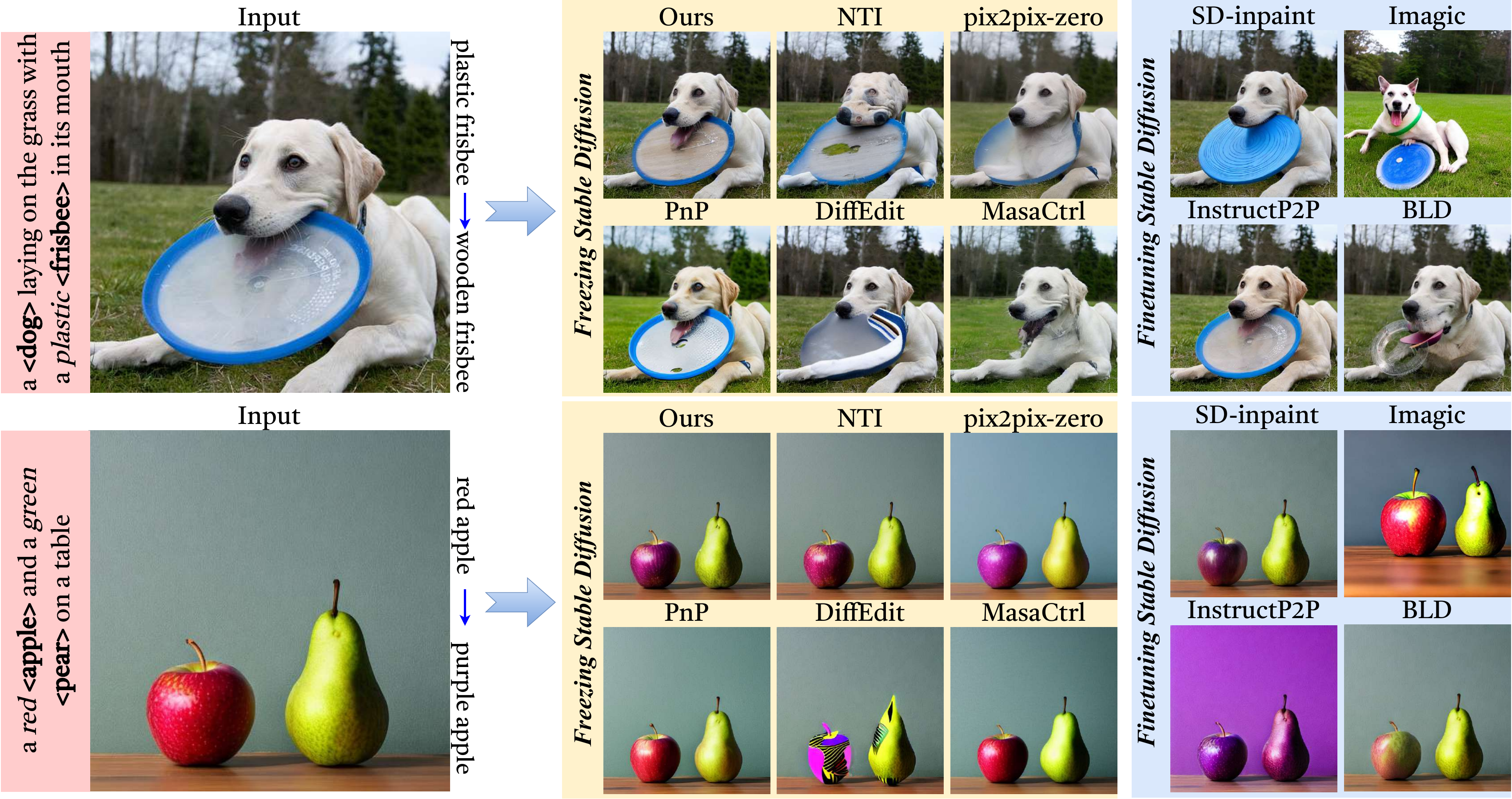}
  \vspace{-3mm}
  \caption{Attribute-Edit by swapping the adjectives given the \textit{Segment-Prior}. By forcing the binding between the cross-attention from the adjective words and corresponding noun words, \ourmethod successfully edits the color or material attribute.}
  \vspace{-2mm}
  \label{fig:attribute_edit}
\end{figure*}

\subsection{Image editing evaluation}
For image editing, we  combine \ourmethod with the P2P~\cite{hertz2022prompt} image editing method. In this paper, we mainly focus on local editing tasks including Word-Swap and Attribute-Edit.

\minisection{Word-Swap.}As shown in Fig.~\ref{fig:local_swap_comp}, we  compare \ourmethod with various methods by swapping one object from the original image given the segmentation maps as localization priors. Our method, \ourmethod, more successfully translates the source object into the target object while maintaining semantic similarities. 
In Table~\ref{tab:eval_metric}, we designed one editing task for each group of images collected in the COCO-Edit dataset (details in the supplementary material). In both single-object and  multi-object editing tasks, \ourmethod achieves better full image evaluation and only performs worse than the inpainting-based methods~\cite{avrahami2023blended,Rombach_2022_CVPR_stablediffusion} in terms of background preservation (since these methods do not change background pixels). It is worth noting that \ourmethod does not require fine-tuning the base model, resulting in better time complexity and no forgetting in the T2I models.
In Table~\ref{tab:user_study}, we question twenty participants to evaluate the image editing performance from two aspects: the editing quality and the background preservation. In both cases, \ourmethod stands out of these six methods of freezing the SD models. Details on the user study are shown in the supplementary.

\minisection{Attribute-Edit.} 
Furthermore, \ourmethod excels in another editing aspect that other methods tend to overlook, which is attribute editing. This capability is illustrated in Fig.~\ref{fig:attribute_edit}. By force the binding between the adjective words and their corresponding noun objects, we achieve the capacity to accurately modify their attributes (color, material, etc.).

\section{Conclusion}
In this paper, we presented \textit{Loc}alization-aware \textit{Inv}ersion (\ourmethod) to solve the cross-attention leakage problem in image editing using text-to-image diffusion models. 
We propose to update the dynamic tokens for each noun word in the prompt with the segmentation or detection as the prior. The resulting cross-attention maps suffer less from cross-attention leakage. Consequently, these greatly improved cross-attention maps result in considerably better results for text-guided image editing. The experimental results, confirm that \ourmethod obtains superior results, especially on complex multi-object scenes. 
Finally, we show that our method can also bind the adjective words to their corresponding nouns, leading to accurate cross-attention maps for the adjectives, and allowing for attribute editing which has not been well explored before for text-guided image editing.

\minisection{Acknowledgments} We acknowledge projects TED2021-132513B-I00 and PID2022-143257NB-I00, financed by MCIN/AEI/10.13039/501100011033 and FSE+ by the European Union NextGenerationEU/PRTR, and by ERDF A Way of Making Europa, and the Generalitat de Catalunya CERCA Program. Chuanming Tang acknowledges the Chinese Scholarship Council (CSC) No.202204910331.

\clearpage

{
    \small
    \bibliographystyle{ieeenat_fullname}
    \bibliography{longstrings,mybib}
}

\clearpage

\appendix
\section{Limitations}
One limitation of our work is related to the size of cross-attention maps. Smaller maps, particularly those sized 16$\times$16, contain more semantic information compared to larger maps. While this rich information is beneficial, it restricts our ability to achieve precise and fine-grained structure control. 
We intend to work on pixel-level text-to-image models to address this limitation.
Another limitation is that the frozen Stable Diffusion (SD) model~\cite{Rombach_2022_CVPR_stablediffusion} itself lacks straightforward editing capabilities, which may impact the quality of editing results. 
To address this limitation, we plan to explore ideas from InstructPix2Pix~\cite{brooks2022instructpix2pix} to develop a better text-guided image editing method in the future.
Additionally, the SD model faces challenges in reconstructing the original image with intricate details due to the compression mechanism in the first-stage autoencoder model. Editing high-frequency information remains a significant and ongoing challenge that requires further research and development.
Overcoming these limitations and advancing our knowledge in these areas will contribute to the improvement and refinement of image editing techniques.
Furthermore, it is noteworthy that diffusion models have shown promise across various applications, such as object detection~\cite{chen2022diffusiondet}, image segmentation~\cite{xu2023odise,pnvr2023ldznet}, and landmark detection~\cite{wu2024diffusion}. Therefore, a potential future research is to extend diffusion models to various practical applications, including visual tracking, continual learning, font generation, and beyond.

\section{Broader Impacts}
The application of text-to-image models in image editing has a wide range of potential uses in various downstream applications. Our model aims to automate and streamline this process, saving time and resources. However, it's important to acknowledge the limitations discussed in this paper. Our model can serve as an intermediate solution, accelerating the creative process and providing insights for future improvements. We must also be aware of potential risks, such as the spread of misinformation, the potential for misuse, and the introduction of biases. Ethical considerations and broader impacts should be carefully examined to responsibly harness the capabilities of these models.

\section{Dataset statistics}

The details of the \textit{COCO-Edit} dataset are presented in Table~\ref{tab:dataset_stat}. In this dataset, we have single-object and multi-object images, with seven and six search prompts, respectively. To assess the image editing quality, each search prompt is associated with one editing task, and we compute the average qualitative metrics for evaluation.

\begin{table*}[t]
    \centering
    \resizebox{0.74\linewidth}{!}{
    \begin{tabular}{c| c|c|c}
      \toprule
      & Prompts &	Image Number & Editing task	 \\
      \midrule
      \multirow{7}{*}{single-object} & airplane	& 25	& airplane $\rightarrow$ seaplane \\
    &bear	& 22	& bear $\rightarrow$ panda\\
        &boat	& 17 & boat $\rightarrow$ canoe\\
    &cat	& 22	& cat $\rightarrow$ cougar\\
        &elephant	& 46& elephant $\rightarrow$ buffalo	\\
   & giraffe	& 37	& giraffe $\rightarrow$ brachiosaurus\\
    &zebra	& 45	& zebra $\rightarrow$ donkey\\
    \midrule
    \multirow{7}{*}{multi-object} & apple, banana	& 2	&apple $\rightarrow$ peach \\
    &bench, cat	& 31	& cat $\rightarrow$ panther\\
    &cat, dog	& 37	& dog $\rightarrow$ fox\\
    &dog, frisbee	& 25	& dog $\rightarrow$ raccon\\
    &donkey, zebra	& 2	& donkey $\rightarrow$ horse\\
    &person, dog, boat	& 4	& dog $\rightarrow$ cat \\
      \bottomrule
    \end{tabular}
    }
    \caption{Comprehensive statistics for the COCO-edit dataset.}
    \label{tab:dataset_stat}
\end{table*}

\section{User Study}

In Table~\ref{tab:supp_user_study}, we provide the full table of our user studies. Each of the twenty participants was randomly assigned twenty image editing tasks from our \textit{COCO-Edit} dataset. Ten tasks involved single-object images, and the other ten involved multi-object images. Participants were asked to anonymously evaluate the best approach out of six methods based on image editing quality and background preservation. The results show that \ourmethod was preferred by the participants, primarily satisfying their subjective preferences for both single-object and multi-object images. For reference, the user study interface is depicted in Fig.~\ref{fig:supp_userstudy}.

\begin{table*}[t]
    \centering
    \resizebox{\linewidth}{!}{
    \begin{tabular}{|c|c|cccccc|}
      \toprule
      \multicolumn{8}{|c|}{User Study (\%)} \\
      \midrule
      & \rotatebox{0}{method}	& \rotatebox{0}{\ourmethod}  &	\rotatebox{0}{DiffEdit~\cite{couairon2023diffedit}}	& \rotatebox{0}{MasaCtrl~\cite{cao2023masactrl}}	& \rotatebox{0}{NTI~\cite{mokady2022null}} & \rotatebox{0}{pix2pix~\cite{parmar2023zero}} 	& \rotatebox{0}{PnP~\cite{tumanyan2022plug}} \\
            \midrule
      \multirow{2}{*}{Single-Object} & Image Editing quality & \textbf{33.5}	& 33.0	& 4.5	& 27.5	& 0.5	& 1.0 \\
      & Background preservation & 20.5	& 4.0	& 5.0	& 22.0	& \textbf{25.0}	& 23.5 \\
            \midrule
      \multirow{2}{*}{Multi-Object} & Image Editing quality & \textbf{46.5}	& 21.0	& 2.5	& 23.5	& 1.0	& 5.5 \\
      & Background preservation & \textbf{31.0}	& 3.5	& 4.0	& 18.0	& 20.5	& 23.0 \\
      
      \midrule
      \multirow{2}{*}{Overall} & Image Editing quality & \textbf{40.0}	& 27.0	& 3.5	& 25.5	& 0.75	& 3.25 \\
      & Background preservation & \textbf{25.8}	& 3.7	& 4.5	& 20.0	& 22.7	& 23.3 \\
      \bottomrule
    \end{tabular}
    }
    \caption{Extended user study with additional statistics over the sing-object and multi-object images separately. Based on the evaluation, it can be inferred that \ourmethod is the best approach, primarily satisfying the evaluators' subjective preferences for both single-object and multi-object images.}
    \label{tab:supp_user_study}
\end{table*}

\section{Cross-Attention Maps}

\begin{figure*}[t]
  \centering
    \includegraphics[width=0.99\textwidth]{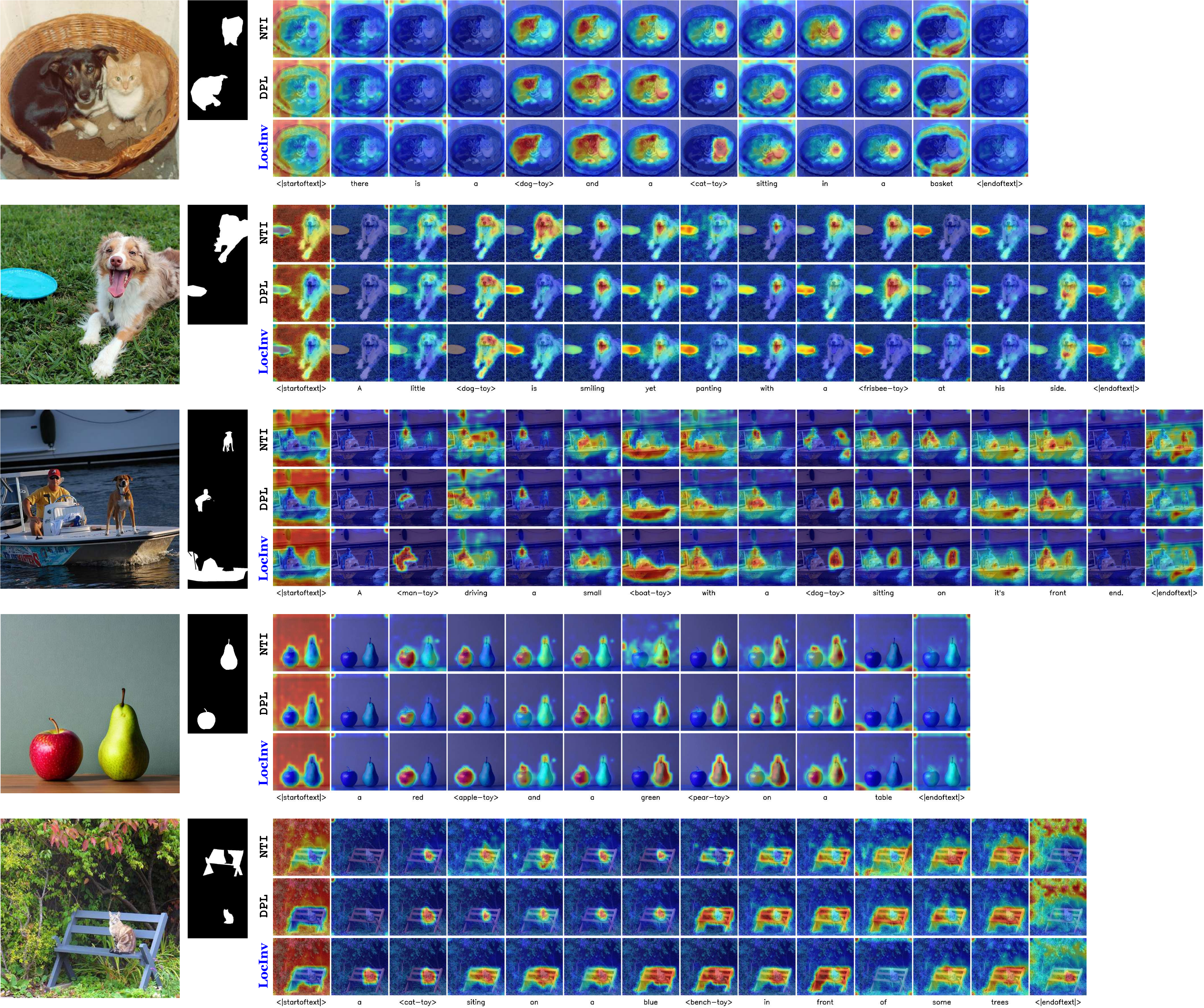}
  \caption{Cross-attention maps are significantly improved by our method, in comparison to previous approaches, including NTI~\cite{mokady2022null}, DDIM~\cite{song2021ddim}, and DPL~\cite{kai2023DPL}. }
  \label{fig:crossattn_change_1}
\end{figure*}

\begin{figure*}[t]
  \centering
    \includegraphics[width=0.99\textwidth]{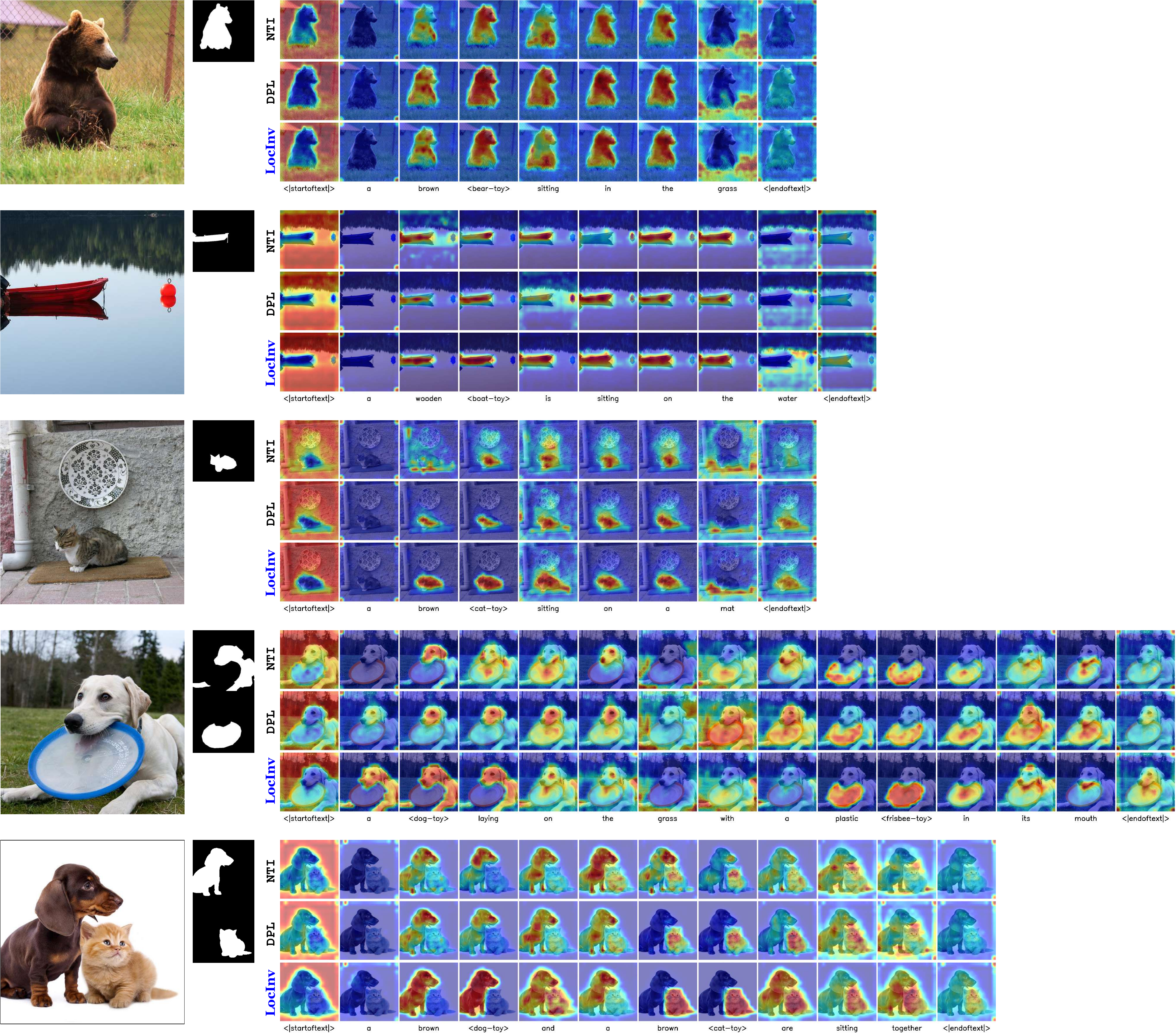}
  \caption{Cross-attention maps are significantly improved by our method, \ourmethod, in comparison to previous approaches, including NTI~\cite{mokady2022null}, DDIM~\cite{song2021ddim}, and DPL~\cite{kai2023DPL}.}
  \label{fig:crossattn_change_2}
\end{figure*}

In addition to the qualitative and quantitative image editing comparisons presented in the main paper, we have extended our evaluation by including a comparison with full sentences cross-attention maps. These comparisons are illustrated in Fig.~\ref{fig:crossattn_change_1} and Fig.~\ref{fig:crossattn_change_2}. 
This comprehensive analysis further demonstrates the superior adaptability and quality of cross-attention maps produced by \ourmethod, in comparison to previous approaches, including NTI~\cite{mokady2022null}, DDIM~\cite{song2021ddim}, and DPL~\cite{kai2023DPL}, even across various input prompt lengths and real image scenarios.

\section{More Image Editing quality comparison}
\subsection{Word-Swap}
\begin{figure*}[t]
  \centering
    \includegraphics[width=0.88\textwidth]{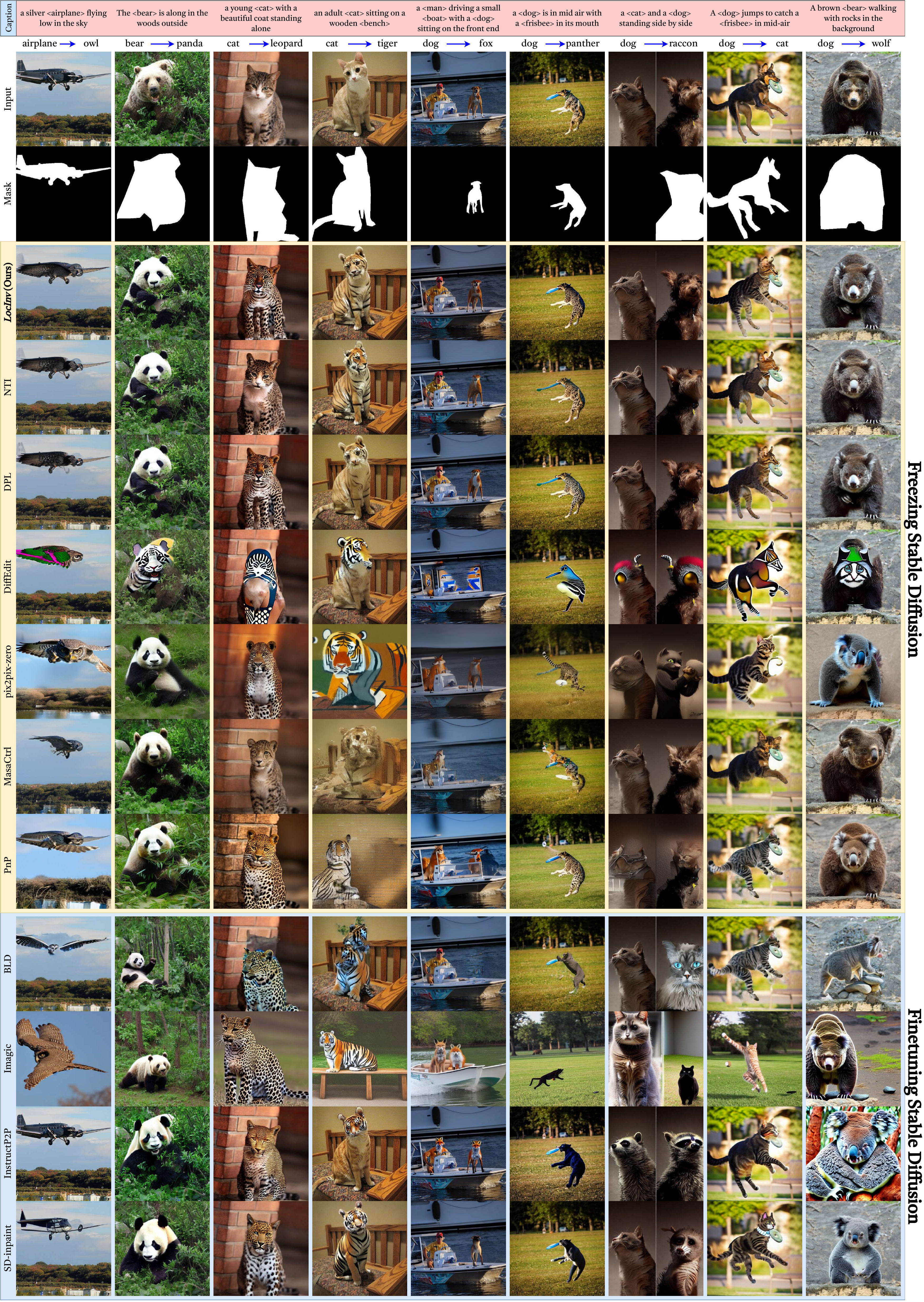}
  \caption{More comparison examples of the local object Word-Swap editing given the \textit{Segment-Prior}. }
  \label{fig:supp_word_swap}
\end{figure*}

In Fig.~\ref{fig:supp_word_swap}, we have provided additional qualitative comparisons of image editing with various methods, including DPL\cite{kai2023DPL}, as an extension of Fig.4 in the main paper. Among these methods, DPL behaves similarly to NTI~\cite{mokady2022null} in complex real images. However, our method, \ourmethod, exhibits superior image editing quality and background preservation in these comparisons.

\subsection{Attribute-Edit}
\begin{figure*}[t]
  \centering
    \includegraphics[width=0.88\textwidth]{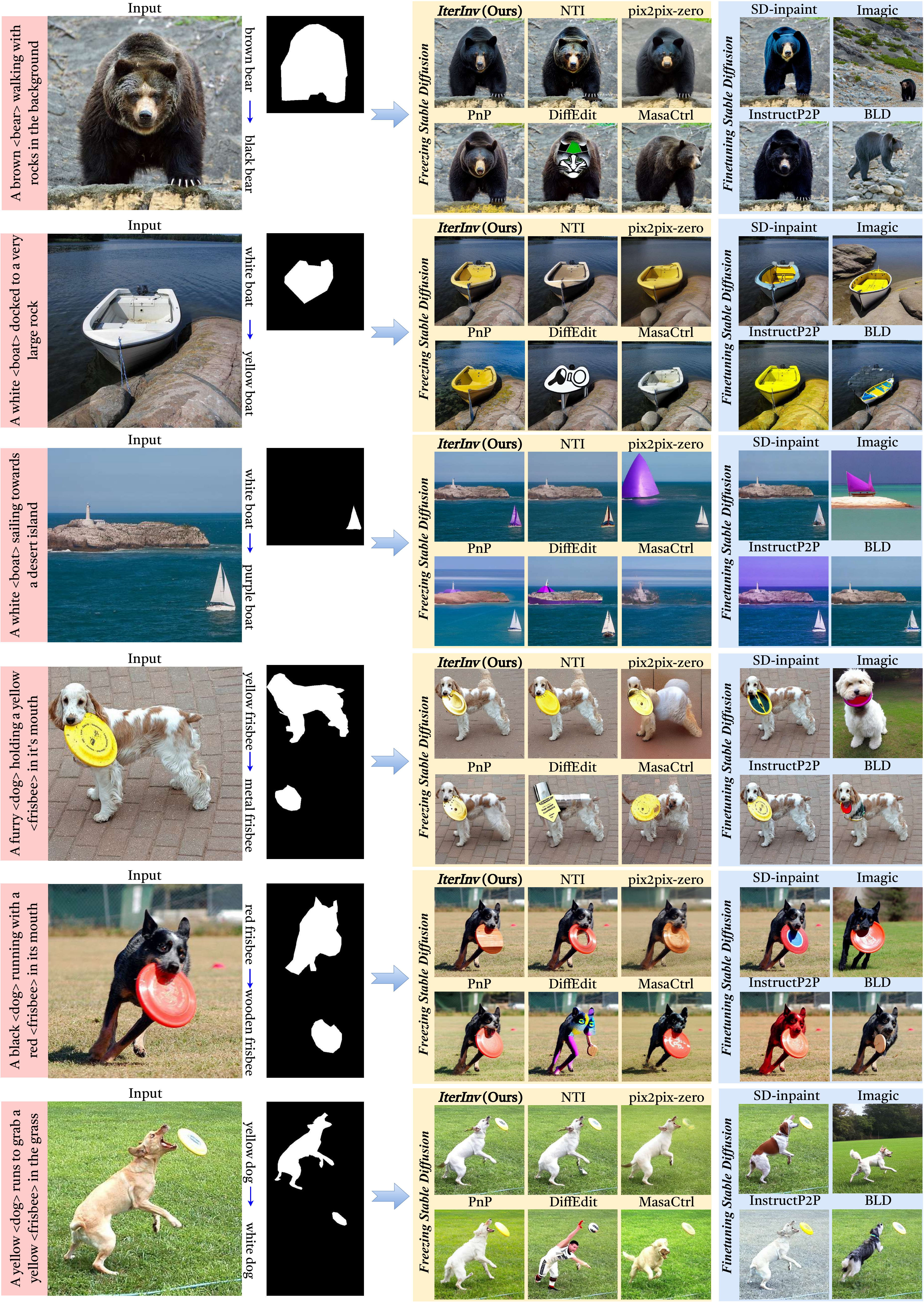}
  \caption{More attribute-Edit with our method \ourmethod by swapping the adjectives given the \textit{Segment-Prior}. 
  }
  \label{fig:supp_attribute_edit}
\end{figure*}

In Fig.~\ref{fig:supp_attribute_edit}, we have presented additional Attribute-Edit tasks as an extension of Fig.5 in the main paper. Once again, our method, \ourmethod, stands out as the best approach, successfully achieving attribute modifications such as color and material with superior results.

\subsection{Between Segment-Prior and Detect-Prior.}
\begin{figure*}[t]
  \centering
    \includegraphics[width=0.999\textwidth]{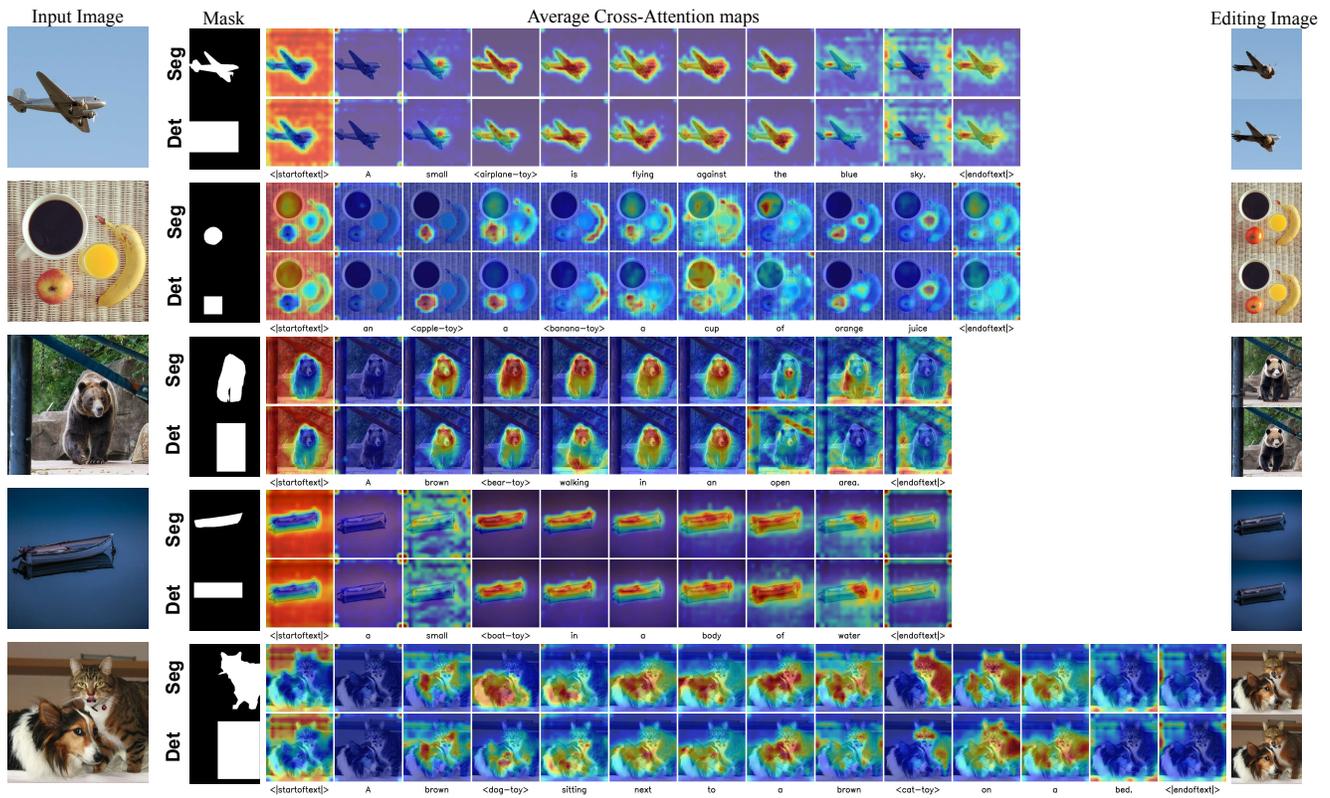}
  \caption{Cross attention maps are better corrected by the Segment-Prior of our method \ourmethod. The “Seg” and “Det” on the left represent the Segment-prior and Detection-Prior, respectively.}
  \label{fig:det_seg_edit}
\end{figure*}

\begin{figure*}[t]
  \centering
    \includegraphics[width=0.85\textwidth]{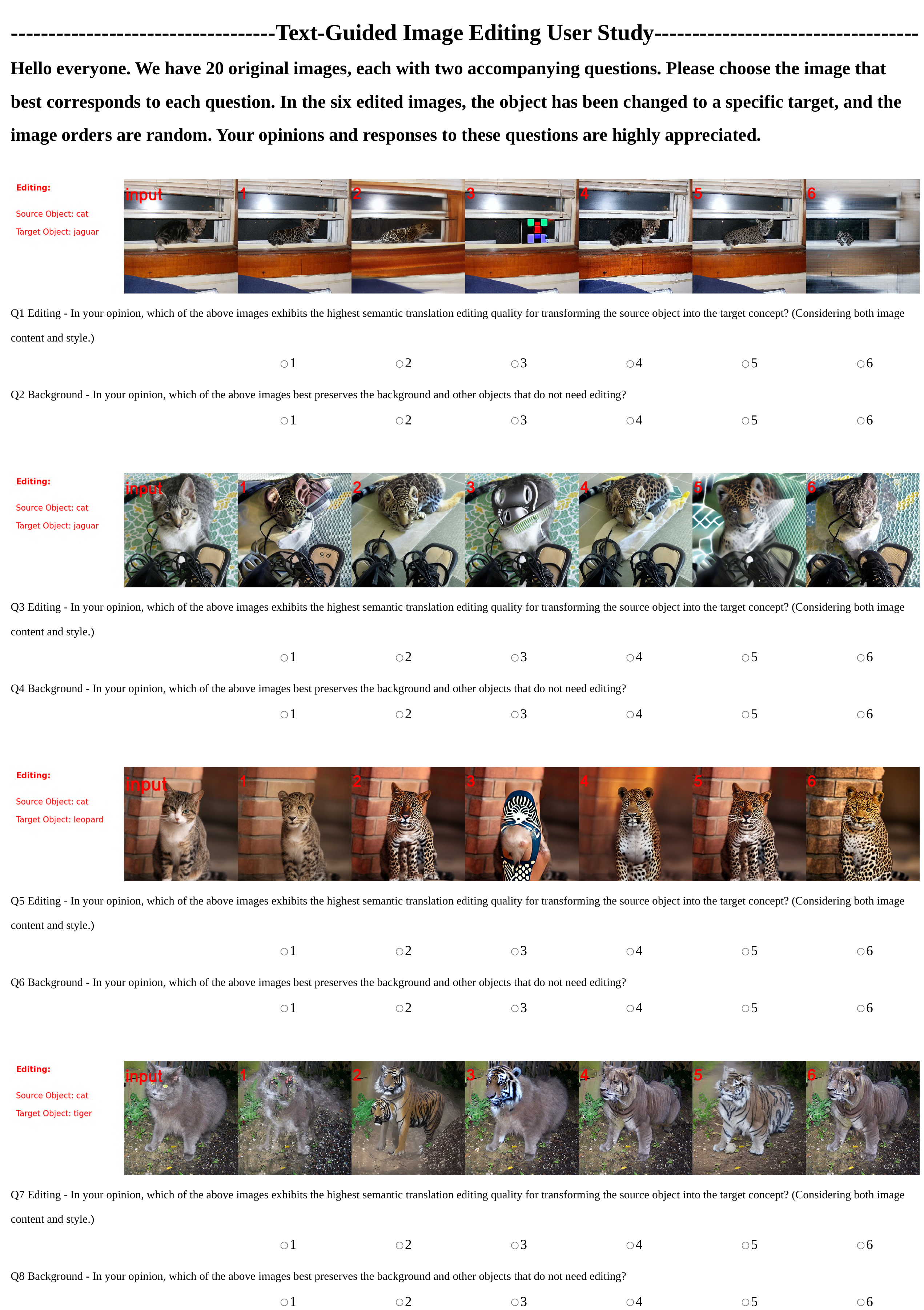}
  \caption{Our user study webpage screenshot. }
  \label{fig:supp_userstudy}
\end{figure*}

In Fig.~\ref{fig:det_seg_edit}, we demonstrate that the Segment-Prior of our method \ourmethod achieves improved alignment between the cross-attention maps and the localization priors, also aligning with our expectations. Consequently, the editing performance varies based on the quality of the cross-attention maps, and Segment-Prior outperforms the Detect-Prior based \ourmethod in this regard.

\end{document}

%% file: preamble.tex
\usepackage[dvipsnames]{xcolor}


%% file: main.bbl
\begin{thebibliography}{65}
\providecommand{\natexlab}[1]{#1}
\providecommand{\url}[1]{\texttt{#1}}
\expandafter\ifx\csname urlstyle\endcsname\relax
  \providecommand{\doi}[1]{doi: #1}\else
  \providecommand{\doi}{doi: \begingroup \urlstyle{rm}\Url}\fi

\bibitem[Avrahami et~al.(2022)Avrahami, Lischinski, and Fried]{avrahami2022blended}
Omri Avrahami, Dani Lischinski, and Ohad Fried.
\newblock Blended diffusion for text-driven editing of natural images.
\newblock In \emph{Proceedings of the IEEE/CVF Conference on Computer Vision and Pattern Recognition}, pages 18208--18218, 2022.

\bibitem[Avrahami et~al.(2023)Avrahami, Fried, and Lischinski]{avrahami2023blended}
Omri Avrahami, Ohad Fried, and Dani Lischinski.
\newblock Blended latent diffusion.
\newblock \emph{ACM Transactions on Graphics (TOG)}, 42\penalty0 (4):\penalty0 1--11, 2023.

\bibitem[Bar-Tal et~al.(2022)Bar-Tal, Ofri-Amar, Fridman, Kasten, and Dekel]{bar2022text2live}
Omer Bar-Tal, Dolev Ofri-Amar, Rafail Fridman, Yoni Kasten, and Tali Dekel.
\newblock Text2live: Text-driven layered image and video editing.
\newblock In \emph{European Conference on Computer Vision}, pages 707--723. Springer, 2022.

\bibitem[Brooks et~al.(2023)Brooks, Holynski, and Efros]{brooks2022instructpix2pix}
Tim Brooks, Aleksander Holynski, and Alexei~A. Efros.
\newblock Instructpix2pix: Learning to follow image editing instructions.
\newblock In \emph{Proceedings of the IEEE Conference on Computer Vision and Pattern Recognition}, 2023.

\bibitem[Cao et~al.(2023)Cao, Wang, Qi, Shan, Qie, and Zheng]{cao2023masactrl}
Mingdeng Cao, Xintao Wang, Zhongang Qi, Ying Shan, Xiaohu Qie, and Yinqiang Zheng.
\newblock Masactrl: Tuning-free mutual self-attention control for consistent image synthesis and editing.
\newblock \emph{Proceedings of the International Conference on Computer Vision}, 2023.

\bibitem[Caron et~al.(2021)Caron, Touvron, Misra, J{\'e}gou, Mairal, Bojanowski, and Joulin]{caron2021dino}
Mathilde Caron, Hugo Touvron, Ishan Misra, Herv{\'e} J{\'e}gou, Julien Mairal, Piotr Bojanowski, and Armand Joulin.
\newblock Emerging properties in self-supervised vision transformers.
\newblock In \emph{Proceedings of the IEEE/CVF international conference on computer vision}, pages 9650--9660, 2021.

\bibitem[Chang et~al.(2023)Chang, Zhang, Barber, Maschinot, Lezama, Jiang, Yang, Murphy, Freeman, Rubinstein, et~al.]{chang2023muse}
Huiwen Chang, Han Zhang, Jarred Barber, AJ Maschinot, Jose Lezama, Lu Jiang, Ming-Hsuan Yang, Kevin Murphy, William~T Freeman, Michael Rubinstein, et~al.
\newblock Muse: Text-to-image generation via masked generative transformers.
\newblock \emph{International Conference on Machine Learning}, 2023.

\bibitem[Chen and Huang(2023)]{chen2023fec}
Songyan Chen and Jiancheng Huang.
\newblock Fec: Three finetuning-free methods to enhance consistency for real image editing.
\newblock \emph{arXiv preprint arXiv:2309.14934}, 2023.

\bibitem[Chen et~al.(2022)Chen, Sun, Song, and Luo]{chen2022diffusiondet}
Shoufa Chen, Peize Sun, Yibing Song, and Ping Luo.
\newblock Diffusiondet: Diffusion model for object detection.
\newblock \emph{arXiv preprint arXiv:2211.09788}, 2022.

\bibitem[Choi et~al.(2021)Choi, Kim, Jeong, Gwon, and Yoon]{choi2021ilvr}
Jooyoung Choi, Sungwon Kim, Yonghyun Jeong, Youngjune Gwon, and Sungroh Yoon.
\newblock Ilvr: Conditioning method for denoising diffusion probabilistic models.
\newblock In \emph{Proceedings of the International Conference on Computer Vision}, pages 14347--14356. IEEE, 2021.

\bibitem[Couairon et~al.(2023)Couairon, Verbeek, Schwenk, and Cord]{couairon2023diffedit}
Guillaume Couairon, Jakob Verbeek, Holger Schwenk, and Matthieu Cord.
\newblock Diffedit: Diffusion-based semantic image editing with mask guidance.
\newblock In \emph{The Eleventh International Conference on Learning Representations}, 2023.

\bibitem[Ding et~al.(2023)Ding, Wang, and Tu]{ding2022maskclip}
Zheng Ding, Jieke Wang, and Zhuowen Tu.
\newblock Open-vocabulary panoptic segmentation with maskclip.
\newblock \emph{International Conference on Machine Learning}, 2023.

\bibitem[Gafni et~al.(2022)Gafni, Polyak, Ashual, Sheynin, Parikh, and Taigman]{gafni2022make}
Oran Gafni, Adam Polyak, Oron Ashual, Shelly Sheynin, Devi Parikh, and Yaniv Taigman.
\newblock Make-a-scene: Scene-based text-to-image generation with human priors.
\newblock In \emph{European Conference on Computer Vision}, pages 89--106. Springer, 2022.

\bibitem[Ghiasi et~al.(2022)Ghiasi, Gu, Cui, and Lin]{ghiasi2022openseg}
Golnaz Ghiasi, Xiuye Gu, Yin Cui, and Tsung-Yi Lin.
\newblock Scaling open-vocabulary image segmentation with image-level labels.
\newblock In \emph{European Conference on Computer Vision}, pages 540--557. Springer, 2022.

\bibitem[Grechka et~al.(2023)Grechka, Couairon, and Cord]{grechka2023gradpaint}
Asya Grechka, Guillaume Couairon, and Matthieu Cord.
\newblock Gradpaint: Gradient-guided inpainting with diffusion models.
\newblock \emph{arXiv preprint arXiv:2309.09614}, 2023.

\bibitem[Han et~al.(2023)Han, Wen, Chen, Zhang, Song, Ren, Gao, Chen, Liu, Zhangli, et~al.]{han2023ProxNPI}
Ligong Han, Song Wen, Qi Chen, Zhixing Zhang, Kunpeng Song, Mengwei Ren, Ruijiang Gao, Yuxiao Chen, Di Liu, Qilong Zhangli, et~al.
\newblock Improving negative-prompt inversion via proximal guidance.
\newblock \emph{arXiv preprint arXiv:2306.05414}, 2023.

\bibitem[Hertz et~al.(2023{\natexlab{a}})Hertz, Aberman, and Cohen-Or]{hertz2023delta_DDS}
Amir Hertz, Kfir Aberman, and Daniel Cohen-Or.
\newblock Delta denoising score.
\newblock \emph{arXiv preprint arXiv:2304.07090}, 2023{\natexlab{a}}.

\bibitem[Hertz et~al.(2023{\natexlab{b}})Hertz, Mokady, Tenenbaum, Aberman, Pritch, and Cohen-Or]{hertz2022prompt}
Amir Hertz, Ron Mokady, Jay Tenenbaum, Kfir Aberman, Yael Pritch, and Daniel Cohen-Or.
\newblock Prompt-to-prompt image editing with cross attention control.
\newblock \emph{International Conference on Learning Representations}, 2023{\natexlab{b}}.

\bibitem[Hessel et~al.(2021)Hessel, Holtzman, Forbes, Le~Bras, and Choi]{hessel2021clipscore}
Jack Hessel, Ari Holtzman, Maxwell Forbes, Ronan Le~Bras, and Yejin Choi.
\newblock Clipscore: A reference-free evaluation metric for image captioning.
\newblock In \emph{Proceedings of the 2021 Conference on Empirical Methods in Natural Language Processing}, pages 7514--7528, 2021.

\bibitem[Ho and Salimans(2022)]{ho2022classifier}
Jonathan Ho and Tim Salimans.
\newblock Classifier-free diffusion guidance.
\newblock \emph{NeurIPS 2021 Workshop on Deep Generative Models and Downstream Applications}, 2022.

\bibitem[Hong et~al.(2023)Hong, Lee, Jang, and Kim]{hong2022sag}
Susung Hong, Gyuseong Lee, Wooseok Jang, and Seungryong Kim.
\newblock Improving sample quality of diffusion models using self-attention guidance.
\newblock \emph{Proceedings of the International Conference on Computer Vision}, 2023.

\bibitem[Honnibal and Montani(2017)]{honnibal2017spacy}
Matthew Honnibal and Ines Montani.
\newblock spacy 2: Natural language understanding with bloom embeddings, convolutional neural networks and incremental parsing.
\newblock \emph{To appear}, 7\penalty0 (1):\penalty0 411--420, 2017.

\bibitem[Huang et~al.(2023)Huang, Liu, Qin, and Chen]{huang2023kv}
Jiancheng Huang, Yifan Liu, Jin Qin, and Shifeng Chen.
\newblock Kv inversion: Kv embeddings learning for text-conditioned real image action editing.
\newblock \emph{arXiv preprint arXiv:2309.16608}, 2023.

\bibitem[Ju et~al.(2023)Ju, Zeng, Bian, Liu, and Xu]{direct_inversion_2023}
Xuan Ju, Ailing Zeng, Yuxuan Bian, Shaoteng Liu, and Qiang Xu.
\newblock Direct inversion: Boosting diffusion-based editing with 3 lines of code.
\newblock \emph{arXiv preprint arXiv:2310.01506}, 2023.

\bibitem[Kawar et~al.(2023)Kawar, Zada, Lang, Tov, Chang, Dekel, Mosseri, and Irani]{kawar2022imagic}
Bahjat Kawar, Shiran Zada, Oran Lang, Omer Tov, Huiwen Chang, Tali Dekel, Inbar Mosseri, and Michal Irani.
\newblock Imagic: Text-based real image editing with diffusion models.
\newblock \emph{Proceedings of the IEEE Conference on Computer Vision and Pattern Recognition}, 2023.

\bibitem[Kim et~al.(2022)Kim, Kwon, and Ye]{Kim_2022_CVPR}
Gwanghyun Kim, Taesung Kwon, and Jong~Chul Ye.
\newblock Diffusionclip: Text-guided diffusion models for robust image manipulation.
\newblock In \emph{Proceedings of the IEEE/CVF Conference on Computer Vision and Pattern Recognition (CVPR)}, pages 2426--2435, 2022.

\bibitem[Kirillov et~al.(2023)Kirillov, Mintun, Ravi, Mao, Rolland, Gustafson, Xiao, Whitehead, Berg, Lo, Doll{\'a}r, and Girshick]{kirillov2023segment_anything_sam}
Alexander Kirillov, Eric Mintun, Nikhila Ravi, Hanzi Mao, Chloe Rolland, Laura Gustafson, Tete Xiao, Spencer Whitehead, Alexander~C. Berg, Wan-Yen Lo, Piotr Doll{\'a}r, and Ross Girshick.
\newblock Segment anything.
\newblock \emph{Proceedings of the International Conference on Computer Vision}, 2023.

\bibitem[Kwon and Ye(2023)]{kwon2023diffusionbased}
Gihyun Kwon and Jong~Chul Ye.
\newblock Diffusion-based image translation using disentangled style and content representation.
\newblock In \emph{The Eleventh International Conference on Learning Representations}, 2023.

\bibitem[Li et~al.(2023)Li, van~de Weijer, Hu, Khan, Hou, Wang, and Yang]{li2023stylediffusion}
Senmao Li, Joost van~de Weijer, Taihang Hu, Fahad~Shahbaz Khan, Qibin Hou, Yaxing Wang, and Jian Yang.
\newblock Stylediffusion: Prompt-embedding inversion for text-based editing, 2023.

\bibitem[Lin et~al.(2014)Lin, Maire, Belongie, Hays, Perona, Ramanan, Doll{\'a}r, and Zitnick]{lin2014coco}
Tsung-Yi Lin, Michael Maire, Serge Belongie, James Hays, Pietro Perona, Deva Ramanan, Piotr Doll{\'a}r, and C~Lawrence Zitnick.
\newblock Microsoft coco: Common objects in context.
\newblock In \emph{European Conference on Computer Vision}, pages 740--755. Springer, 2014.

\bibitem[Liu et~al.(2023)Liu, Zeng, Ren, Li, Zhang, Yang, Li, Yang, Su, Zhu, and Zhang]{ShilongLiu2023GroundingDino}
Shilong Liu, Zhaoyang Zeng, Tianhe Ren, Feng Li, Hao Zhang, Jie Yang, Chunyuan Li, Jianwei Yang, Hang Su, Jun Zhu, and Lei Zhang.
\newblock Grounding dino: Marrying dino with grounded pre-training for open-set object detection.
\newblock In \emph{arXiv preprint arXiv:2303.05499}, 2023.

\bibitem[L{\"u}ddecke and Ecker(2022)]{luddecke2022clipseg}
Timo L{\"u}ddecke and Alexander Ecker.
\newblock Image segmentation using text and image prompts.
\newblock In \emph{Proceedings of the IEEE/CVF Conference on Computer Vision and Pattern Recognition}, pages 7086--7096, 2022.

\bibitem[Lugmayr et~al.(2022)Lugmayr, Danelljan, Romero, Yu, Timofte, and Van~Gool]{lugmayr2022repaint}
Andreas Lugmayr, Martin Danelljan, Andres Romero, Fisher Yu, Radu Timofte, and Luc Van~Gool.
\newblock Repaint: Inpainting using denoising diffusion probabilistic models.
\newblock In \emph{Proceedings of the IEEE/CVF Conference on Computer Vision and Pattern Recognition}, pages 11461--11471, 2022.

\bibitem[Meng et~al.(2022)Meng, He, Song, Song, Wu, Zhu, and Ermon]{meng2022sdedit}
Chenlin Meng, Yutong He, Yang Song, Jiaming Song, Jiajun Wu, Jun-Yan Zhu, and Stefano Ermon.
\newblock {SDE}dit: Guided image synthesis and editing with stochastic differential equations.
\newblock In \emph{International Conference on Learning Representations}, 2022.

\bibitem[{Midjourney.com}(2022)]{midjourney}
{Midjourney.com}.
\newblock Midjourney.
\newblock \url{https://www.midjourney.com}, 2022.

\bibitem[Miyake et~al.(2023)Miyake, Iohara, Saito, and Tanaka]{miyake2023NPI}
Daiki Miyake, Akihiro Iohara, Yu Saito, and Toshiyuki Tanaka.
\newblock Negative-prompt inversion: Fast image inversion for editing with text-guided diffusion models.
\newblock \emph{arXiv preprint arXiv:2305.16807}, 2023.

\bibitem[Mokady et~al.(2023)Mokady, Hertz, Aberman, Pritch, and Cohen-Or]{mokady2022null}
Ron Mokady, Amir Hertz, Kfir Aberman, Yael Pritch, and Daniel Cohen-Or.
\newblock Null-text inversion for editing real images using guided diffusion models.
\newblock \emph{Proceedings of the IEEE Conference on Computer Vision and Pattern Recognition}, 2023.

\bibitem[Nichol et~al.(2022)Nichol, Dhariwal, Ramesh, Shyam, Mishkin, Mcgrew, Sutskever, and Chen]{GLIDE}
Alexander~Quinn Nichol, Prafulla Dhariwal, Aditya Ramesh, Pranav Shyam, Pamela Mishkin, Bob Mcgrew, Ilya Sutskever, and Mark Chen.
\newblock {GLIDE}: Towards photorealistic image generation and editing with text-guided diffusion models.
\newblock In \emph{Proceedings of the 39th International Conference on Machine Learning}, pages 16784--16804. PMLR, 2022.

\bibitem[Parmar et~al.(2023)Parmar, Singh, Zhang, Li, Lu, and Zhu]{parmar2023zero}
Gaurav Parmar, Krishna~Kumar Singh, Richard Zhang, Yijun Li, Jingwan Lu, and Jun-Yan Zhu.
\newblock Zero-shot image-to-image translation.
\newblock \emph{Proceedings of the ACM SIGGRAPH Conference on Computer Graphics}, 2023.

\bibitem[Patashnik et~al.(2023)Patashnik, Garibi, Azuri, Averbuch-Elor, and Cohen-Or]{patashnik2023localizing}
Or Patashnik, Daniel Garibi, Idan Azuri, Hadar Averbuch-Elor, and Daniel Cohen-Or.
\newblock Localizing object-level shape variations with text-to-image diffusion models.
\newblock \emph{Proceedings of the International Conference on Computer Vision}, 2023.

\bibitem[Pnvr et~al.(2023)Pnvr, Singh, Ghosh, Siddiquie, and Jacobs]{pnvr2023ldznet}
Koutilya Pnvr, Bharat Singh, Pallabi Ghosh, Behjat Siddiquie, and David Jacobs.
\newblock Ld-znet: A latent diffusion approach for text-based image segmentation.
\newblock In \emph{Proceedings of the IEEE/CVF International Conference on Computer Vision}, pages 4157--4168, 2023.

\bibitem[Podell et~al.(2023)Podell, English, Lacey, Blattmann, Dockhorn, M{\"u}ller, Penna, and Rombach]{podell2023sdxl}
Dustin Podell, Zion English, Kyle Lacey, Andreas Blattmann, Tim Dockhorn, Jonas M{\"u}ller, Joe Penna, and Robin Rombach.
\newblock Sdxl: improving latent diffusion models for high-resolution image synthesis.
\newblock \emph{arXiv preprint arXiv:2307.01952}, 2023.

\bibitem[Ramesh et~al.(2021)Ramesh, Pavlov, Goh, Gray, Voss, Radford, Chen, and Sutskever]{ramesh2021zero}
Aditya Ramesh, Mikhail Pavlov, Gabriel Goh, Scott Gray, Chelsea Voss, Alec Radford, Mark Chen, and Ilya Sutskever.
\newblock Zero-shot text-to-image generation.
\newblock In \emph{International Conference on Machine Learning}, pages 8821--8831. PMLR, 2021.

\bibitem[Ramesh et~al.(2022)Ramesh, Dhariwal, Nichol, Chu, and Chen]{ramesh2022dalle2}
Aditya Ramesh, Prafulla Dhariwal, Alex Nichol, Casey Chu, and Mark Chen.
\newblock Hierarchical text-conditional image generation with clip latents.
\newblock \emph{arXiv preprint arXiv:2204.06125}, 2022.

\bibitem[Rombach et~al.(2022)Rombach, Blattmann, Lorenz, Esser, and Ommer]{Rombach_2022_CVPR_stablediffusion}
Robin Rombach, Andreas Blattmann, Dominik Lorenz, Patrick Esser, and Bj\"orn Ommer.
\newblock High-resolution image synthesis with latent diffusion models.
\newblock In \emph{Proceedings of the IEEE/CVF Conference on Computer Vision and Pattern Recognition (CVPR)}, pages 10684--10695, 2022.

\bibitem[Ronneberger et~al.(2015)Ronneberger, Fischer, and Brox]{ronneberger2015unet}
Olaf Ronneberger, Philipp Fischer, and Thomas Brox.
\newblock U-net: Convolutional networks for biomedical image segmentation.
\newblock In \emph{Medical Image Computing and Computer-Assisted Intervention--MICCAI 2015: 18th International Conference, Munich, Germany, October 5-9, 2015, Proceedings, Part III 18}, pages 234--241. Springer, 2015.

\bibitem[Saharia et~al.(2022)Saharia, Chan, Saxena, Li, Whang, Denton, Ghasemipour, Ayan, Mahdavi, Lopes, et~al.]{saharia2022imagen}
Chitwan Saharia, William Chan, Saurabh Saxena, Lala Li, Jay Whang, Emily Denton, Seyed Kamyar~Seyed Ghasemipour, Burcu~Karagol Ayan, S~Sara Mahdavi, Rapha~Gontijo Lopes, et~al.
\newblock Photorealistic text-to-image diffusion models with deep language understanding.
\newblock \emph{Advances in Neural Information Processing Systems}, 2022.

\bibitem[Shonenkov et~al.(2023)Shonenkov, Konstantinov, Bakshandaeva, Schuhmann, Ivanova, and Klokova]{deepfloyd}
Alex Shonenkov, Misha Konstantinov, Daria Bakshandaeva, Christoph Schuhmann, Ksenia Ivanova, and Nadiia Klokova.
\newblock Deepfloyd-if.
\newblock \url{https://github.com/deep-floyd/IF}, 2023.

\bibitem[Song et~al.(2021)Song, Meng, and Ermon]{song2021ddim}
Jiaming Song, Chenlin Meng, and Stefano Ermon.
\newblock Denoising diffusion implicit models.
\newblock In \emph{International Conference on Learning Representations}, 2021.

\bibitem[Tang et~al.(2023)Tang, Wang, and van~de Weijer]{tang2023iterinv}
Chuanming Tang, Kai Wang, and Joost van~de Weijer.
\newblock Iterinv: Iterative inversion for pixel-level t2i models.
\newblock \emph{Neurips 2023 workshop on Diffusion Models}, 2023.

\bibitem[Tumanyan et~al.(2022)Tumanyan, Bar-Tal, Bagon, and Dekel]{tumanyan2022structure}
Narek Tumanyan, Omer Bar-Tal, Shai Bagon, and Tali Dekel.
\newblock Splicing vit features for semantic appearance transfer.
\newblock In \emph{Proceedings of the IEEE/CVF Conference on Computer Vision and Pattern Recognition}, pages 10748--10757, 2022.

\bibitem[Tumanyan et~al.(2023)Tumanyan, Geyer, Bagon, and Dekel]{tumanyan2022plug}
Narek Tumanyan, Michal Geyer, Shai Bagon, and Tali Dekel.
\newblock Plug-and-play diffusion features for text-driven image-to-image translation.
\newblock \emph{Proceedings of the IEEE Conference on Computer Vision and Pattern Recognition}, 2023.

\bibitem[Wang et~al.(2023{\natexlab{a}})Wang, Yang, Yang, Butt, and van~de Weijer]{kai2023DPL}
Kai Wang, Fei Yang, Shiqi Yang, Muhammad~Atif Butt, and Joost van~de Weijer.
\newblock Dynamic prompt learning: Addressing cross-attention leakage for text-based image editing.
\newblock \emph{Advances in Neural Information Processing Systems}, 2023{\natexlab{a}}.

\bibitem[Wang et~al.(2023{\natexlab{b}})Wang, Zhang, Birsak, and Wonka]{wang2023mdp}
Qian Wang, Biao Zhang, Michael Birsak, and Peter Wonka.
\newblock Mdp: A generalized framework for text-guided image editing by manipulating the diffusion path, 2023{\natexlab{b}}.

\bibitem[Wang et~al.(2003)Wang, Simoncelli, and Bovik]{wang2003ssim}
Zhou Wang, Eero~P Simoncelli, and Alan~C Bovik.
\newblock Multiscale structural similarity for image quality assessment.
\newblock In \emph{The Thrity-Seventh Asilomar Conference on Signals, Systems \& Computers, 2003}, pages 1398--1402. Ieee, 2003.

\bibitem[Wang et~al.(2023{\natexlab{c}})Wang, Li, Chen, Lim, Torralba, Zhao, and Wang]{wang2023unidetector}
Zhenyu Wang, Yali Li, Xi Chen, Ser-Nam Lim, Antonio Torralba, Hengshuang Zhao, and Shengjin Wang.
\newblock Detecting everything in the open world: Towards universal object detection.
\newblock In \emph{Proceedings of the IEEE/CVF Conference on Computer Vision and Pattern Recognition}, pages 11433--11443, 2023{\natexlab{c}}.

\bibitem[Wu et~al.(2024)Wu, Wang, Tang, and Zhang]{wu2024diffusion}
Tao Wu, Kai Wang, Chuanming Tang, and Jianlin Zhang.
\newblock Diffusion-based network for unsupervised landmark detection.
\newblock \emph{Knowledge-Based Systems}, page 111627, 2024.

\bibitem[Xu et~al.(2022)Xu, De~Mello, Liu, Byeon, Breuel, Kautz, and Wang]{xu2022groupvit}
Jiarui Xu, Shalini De~Mello, Sifei Liu, Wonmin Byeon, Thomas Breuel, Jan Kautz, and Xiaolong Wang.
\newblock Groupvit: Semantic segmentation emerges from text supervision.
\newblock In \emph{Proceedings of the IEEE/CVF Conference on Computer Vision and Pattern Recognition}, pages 18134--18144, 2022.

\bibitem[Xu et~al.(2023)Xu, Liu, Vahdat, Byeon, Wang, and De~Mello]{xu2023odise}
Jiarui Xu, Sifei Liu, Arash Vahdat, Wonmin Byeon, Xiaolong Wang, and Shalini De~Mello.
\newblock Open-vocabulary panoptic segmentation with text-to-image diffusion models.
\newblock In \emph{Proceedings of the IEEE/CVF Conference on Computer Vision and Pattern Recognition}, pages 2955--2966, 2023.

\bibitem[Yu et~al.(2023)Yu, Feng, Feng, Liu, Jin, Zeng, and Chen]{yu2023inpaint_anything}
Tao Yu, Runseng Feng, Ruoyu Feng, Jinming Liu, Xin Jin, Wenjun Zeng, and Zhibo Chen.
\newblock Inpaint anything: Segment anything meets image inpainting.
\newblock \emph{arXiv preprint arXiv:2304.06790}, 2023.

\bibitem[Zhang et~al.(2018)Zhang, Isola, Efros, Shechtman, and Wang]{zhang2018lpips}
Richard Zhang, Phillip Isola, Alexei~A Efros, Eli Shechtman, and Oliver Wang.
\newblock The unreasonable effectiveness of deep features as a perceptual metric.
\newblock In \emph{Proceedings of the IEEE conference on computer vision and pattern recognition}, pages 586--595, 2018.

\bibitem[Zhang et~al.(2023)Zhang, Xiao, and Huang]{zhang2023forgedit}
Shiwen Zhang, Shuai Xiao, and Weilin Huang.
\newblock Forgedit: Text guided image editing via learning and forgetting.
\newblock \emph{arXiv preprint arXiv:2309.10556}, 2023.

\bibitem[Zhou et~al.(2023)Zhou, Liu, Zhu, Yang, Chen, and Xu]{zhou2023shifted}
Yufan Zhou, Bingchen Liu, Yizhe Zhu, Xiao Yang, Changyou Chen, and Jinhui Xu.
\newblock Shifted diffusion for text-to-image generation.
\newblock In \emph{Proceedings of the IEEE/CVF Conference on Computer Vision and Pattern Recognition}, pages 10157--10166, 2023.

\bibitem[Zou et~al.(2023{\natexlab{a}})Zou, Dou, Yang, Gan, Li, Li, Dai, Behl, Wang, Yuan, et~al.]{zou2023xdecoder}
Xueyan Zou, Zi-Yi Dou, Jianwei Yang, Zhe Gan, Linjie Li, Chunyuan Li, Xiyang Dai, Harkirat Behl, Jianfeng Wang, Lu Yuan, et~al.
\newblock Generalized decoding for pixel, image, and language.
\newblock In \emph{Proceedings of the IEEE/CVF Conference on Computer Vision and Pattern Recognition}, pages 15116--15127, 2023{\natexlab{a}}.

\bibitem[Zou et~al.(2023{\natexlab{b}})Zou, Yang, Zhang, Li, Li, Gao, and Lee]{zou2023seem}
Xueyan Zou, Jianwei Yang, Hao Zhang, Feng Li, Linjie Li, Jianfeng Gao, and Yong~Jae Lee.
\newblock Segment everything everywhere all at once.
\newblock \emph{Advances in Neural Information Processing Systems}, 2023{\natexlab{b}}.

\end{thebibliography}
